\documentclass[conference]{IEEEtran}

\usepackage{times}
\usepackage{epsfig}
\usepackage{graphicx}
\usepackage{float}
\usepackage{wrapfig}

\usepackage{amsmath,amssymb}

\usepackage{bm,xspace}
\usepackage{comment}
\usepackage{verbatim}
\usepackage{multirow}
\usepackage{balance}
\usepackage{url}
\usepackage{booktabs}
\usepackage{etoolbox,siunitx}
\usepackage{calc}
\usepackage{pifont,hologo}
\usepackage{color}

\newcommand{\ra}[1]{\renewcommand{\arraystretch}{#1}}

\usepackage[super]{nth}
\usepackage{nicefrac}
\def\tt{\mathbf{t}}

\DeclareMathSymbol{@}{\mathord}{letters}{"3B}

\newcommand\mypara[1]{\vspace{3pt}\noindent\textbf{#1.}}

\def\latex/{\LaTeX}
\def\bibtex/{\hologo{BibTeX}}

\usepackage[official]{eurosym}

\usepackage{times}

\usepackage{amsmath,amssymb,amsfonts}
\usepackage{balance}
\usepackage{bm}
\usepackage{hyperref}
\usepackage{color}
\usepackage[skip=0pt,font=footnotesize]{caption}
\usepackage[table,xcdraw]{xcolor}
\usepackage{tabularx}
\usepackage{booktabs}
\usepackage{graphicx}
\usepackage{multicol}
\usepackage{multirow}
\usepackage[numbers, square, comma, sort&compress]{natbib}
\usepackage{pstool}
\usepackage{tikz}
\usepackage{tikz-3dplot}
\usepackage{pgfplots}
\usetikzlibrary{bending, external}

\usepackage[normalem]{ulem}

\newcommand{\wfr}[0]{\ensuremath{\mathcal{W}}} %
\newcommand{\bfr}[0]{\ensuremath{\mathcal{B}}} %
\newcommand{\pfr}[0]{\ensuremath{\mathcal{P}}}

\newcommand{\rotateRPY}[3]%
{   \pgfmathsetmacro{\rollangle}{#1}
    \pgfmathsetmacro{\pitchangle}{#2}
    \pgfmathsetmacro{\yawangle}{#3}

    \pgfmathsetmacro{\newxx}{cos(\yawangle)*cos(\pitchangle)}
    \pgfmathsetmacro{\newxy}{sin(\yawangle)*cos(\pitchangle)}
    \pgfmathsetmacro{\newxz}{-sin(\pitchangle)}
    \path (\newxx,\newxy,\newxz);
    \pgfgetlastxy{\nxx}{\nxy};

    \pgfmathsetmacro{\newyx}{cos(\yawangle)*sin(\pitchangle)*sin(\rollangle)-sin(\yawangle)*cos(\rollangle)}
    \pgfmathsetmacro{\newyy}{sin(\yawangle)*sin(\pitchangle)*sin(\rollangle)+ cos(\yawangle)*cos(\rollangle)}
    \pgfmathsetmacro{\newyz}{cos(\pitchangle)*sin(\rollangle)}
    \path (\newyx,\newyy,\newyz);
    \pgfgetlastxy{\nyx}{\nyy};

    \pgfmathsetmacro{\newzx}{cos(\yawangle)*sin(\pitchangle)*cos(\rollangle)+ sin(\yawangle)*sin(\rollangle)}
    \pgfmathsetmacro{\newzy}{sin(\yawangle)*sin(\pitchangle)*cos(\rollangle)-cos(\yawangle)*sin(\rollangle)}
    \pgfmathsetmacro{\newzz}{cos(\pitchangle)*cos(\rollangle)}
    \path (\newzx,\newzy,\newzz);
    \pgfgetlastxy{\nzx}{\nzy};
}

\newcolumntype{C}{>{\centering\arraybackslash}X}

\newcommand\clearrow{\global\let\rowmac\relax}
\clearrow

\renewcommand{\i}{\text{i}}

\newcommand{\mat}[1]{\begin{bmatrix} #1 \end{bmatrix}}

\newcommand{\rom}[1]{\expandafter{\romannumeral #1\relax}}

\newcommand{\hide}[1]{}

\newcommand{\rev}[1]{{\color{black}#1}}

\definecolor{c1}{HTML}{D9B08C}
\definecolor{c2}{HTML}{FFCB9A}
\definecolor{c3}{HTML}{D1E8FF}
\definecolor{c4}{HTML}{DEF2F1}

\definecolor{somegray}{rgb}{0.5, 0.5, 0.5}
\newcommand{\darkgrayed}[1]{\textcolor{somegray}{#1}}
\makeatletter
\newcommand*\titleheader[1]{\gdef\@titleheader{#1}}
\AtBeginDocument{%
  \let\st@red@title\@title
  \def\@title{%
    \vskip-2em
    \bgroup\normalfont\large\centering\@titleheader\par\egroup
    \vskip1.1em\st@red@title}
}
\makeatother

\titleheader{\darkgrayed{This paper has been accepted for publication at Robotics: Science and Systems 2021 conference.}}

\pgfplotsset{compat=newest}

\IEEEoverridecommandlockouts

\title{NeuroBEM: Hybrid Aerodynamic Quadrotor Model}
\begin{document}

\author{Leonard Bauersfeld$^*$, Elia Kaufmann$^*$, Philipp Foehn, Sihao Sun, Davide Scaramuzza}

\makeatletter
\g@addto@macro\@maketitle{
  \captionsetup{type=figure}\setcounter{figure}{0}
  \def\mycolspace{1.2mm}
 	\begin{tabular}{@{}c@{\hspace{\mycolspace}}c@{\hspace{\mycolspace}}c@{}}
\includegraphics[width=0.34\textwidth,trim=180 260 190 100, clip]{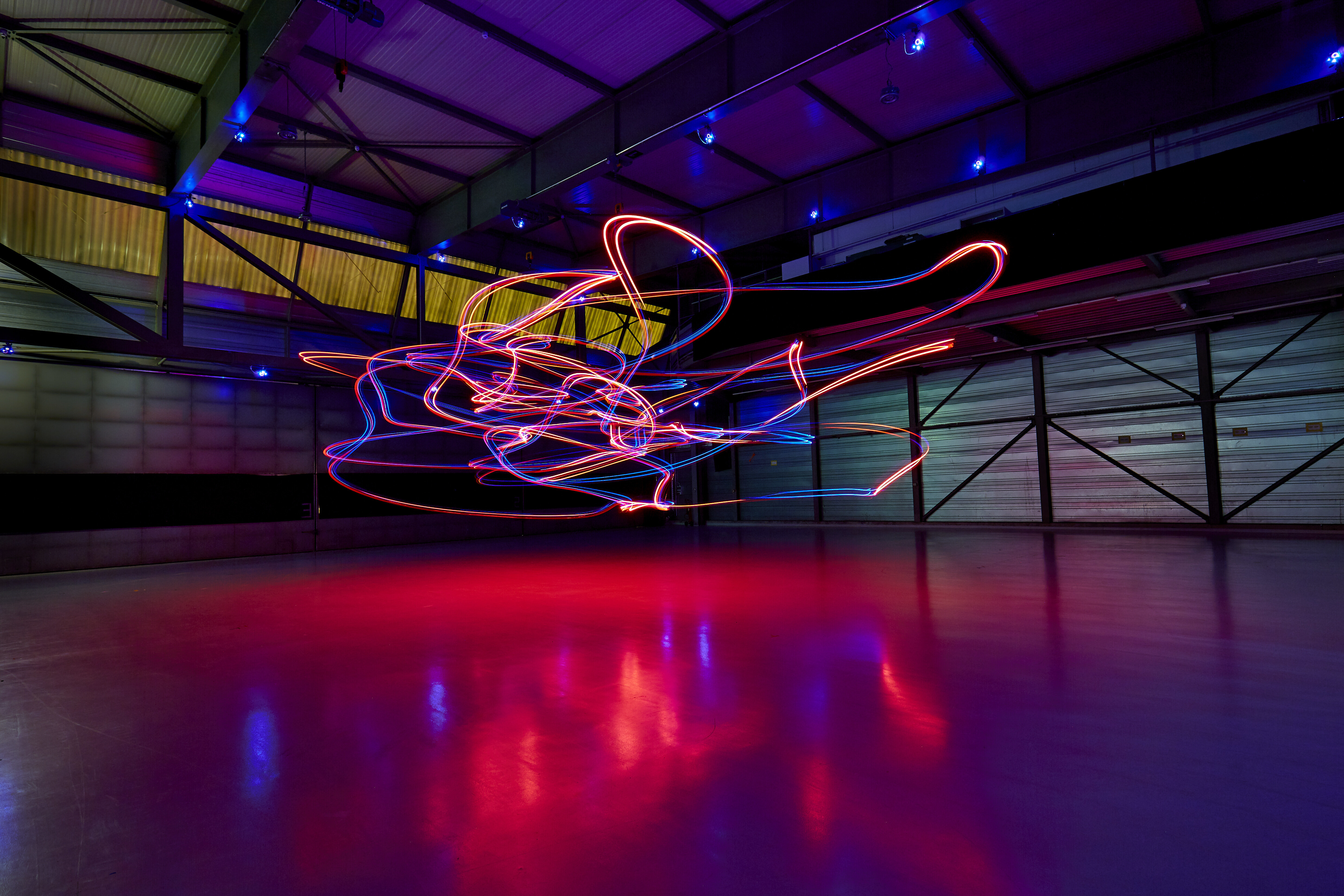} &
\includegraphics[width=0.275\textwidth,trim=70 49 70 50,clip]{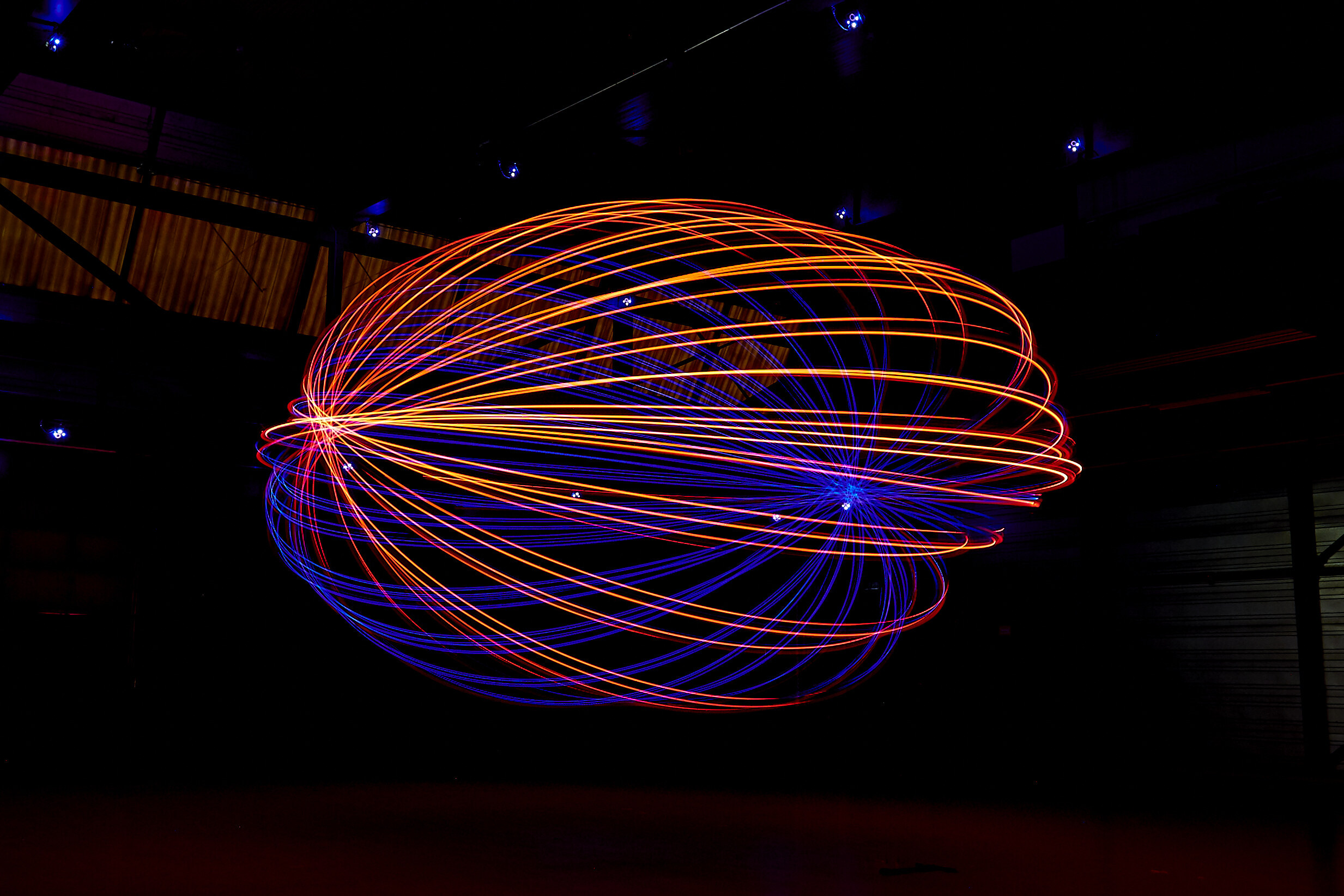} &
\includegraphics[width=0.37\textwidth,trim=0 0 0 0,clip]{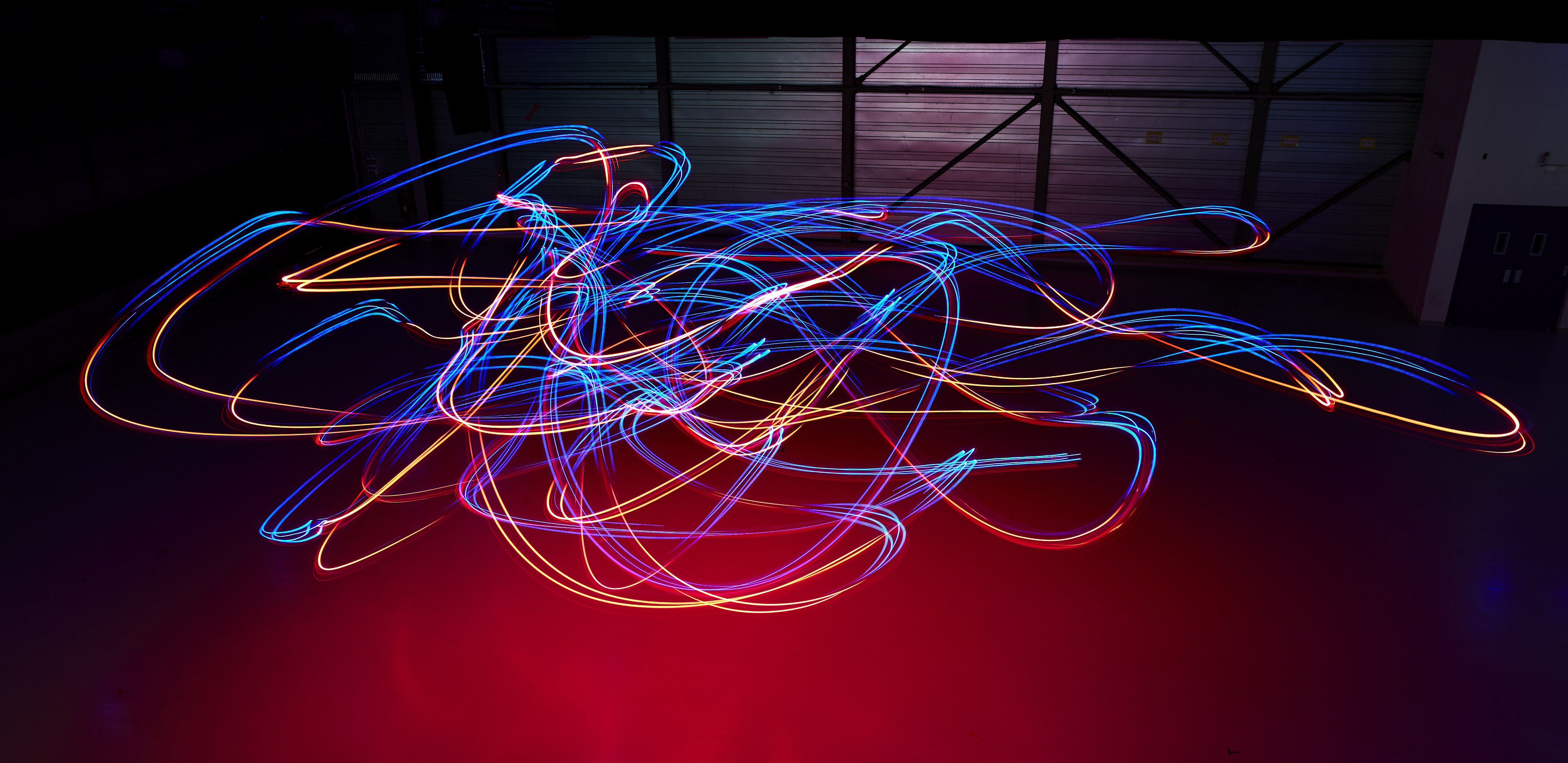}
	\end{tabular}
	\captionof{figure}{Long-exposure images depicting quadrotor trajectory tracking at speeds up to \SI{65} km/h in a large-scale motion-capture system.
    The captured data is used to fit a hybrid  quadrotor model combining blade-element-momentum (BEM) theory with a neural network compensating residual dynamics. This hybrid model reproduces the flown trajectories in simulation with a positional RMSE error reduction of \rev{over 50\%} compared to state-of-the-art.
	\label{fig:catcheye}}
	\vspace{-9pt}
	\thanks{\\[-16pt]$^*$ Equal contribution \\The authors are with the Robotics and Perception Group, Dep. of Informatics, University of Zurich, and Dep. of Neuroinformatics, University of Zurich
and ETH Zurich, Switzerland (\url{http://rpg.ifi.uzh.ch}).
}
}
\makeatother

\maketitle
\begin{abstract}
Quadrotors are extremely agile, 
so much in fact, that classic first-principle-models come to their limits.
Aerodynamic effects, while insignificant at low speeds, become the dominant model defect during high speeds or agile maneuvers.
Accurate modeling is needed to design robust high-performance control systems and enable flying close to the platform's physical limits.
We propose a hybrid approach fusing first principles and learning to model quadrotors and their aerodynamic effects with unprecedented accuracy.
First principles fail to capture such aerodynamic effects, rendering traditional approaches inaccurate when used for simulation or controller tuning.
Data-driven approaches try to capture aerodynamic effects with blackbox modeling, such as neural networks; however, they struggle to robustly generalize to arbitrary flight conditions.
Our hybrid approach unifies \emph{and outperforms} both first-principles blade-element momentum theory and learned residual dynamics.
It is evaluated in one of the world's largest motion-capture systems, using autonomous-quadrotor-flight data at speeds up to 65\,km/h.
The resulting model captures the aerodynamic thrust, torques, and parasitic effects with astonishing accuracy, outperforming existing models with 50\% reduced prediction errors, and shows strong generalization capabilities beyond the training set.

\end{abstract}

\IEEEpeerreviewmaketitle

\vspace*{-3pt}
\section*{Supplementary Material}
\vspace*{-3pt}
A narrated video illustrating our approach is available at {\url{https://youtu.be/Nze1wlfmzTQ}}.
Code and dataset can be found at \url{http://rpg.ifi.uzh.ch/NeuroBEM.html}.
\vspace*{-3pt}
\section{Introduction}\label{sec:introduction}
\vspace*{-3pt}
\setcounter{figure}{1}
In recent years, research on fast navigation of autonomous quadrotors has made tremendous progress, continually pushing the vehicles to more aggressive maneuvers~\cite{ryou2020multi, loianno2017estimation, kaufmann2020RSS, foehn2021time} (Figure \ref{fig:catcheye}).
To further advance the field, several competitions have been organized, such as the autonomous drone race series at the recent IROS and NeurIPS conferences~\cite{moon2019challenges,madaan2019gameofdrones} and the AlphaPilot challenge~\cite{foehn2020alphapilot}.
In the near future, estimation and control algorithms will reach the level of maturity necessary to push autonomous quadrotors to the bounds of what is physically possible. %
This presents the need for quadrotor models that can predict the behaviour of the platform even during highly aggressive maneuvers.

Accurately modeling quadrotors flying at their physical limits is extremely challenging and requires to capture complex effects due to aerodynamic forces, motor dynamics, and vibrations. 
Especially aerodynamic forces pose a challenge, as they depend on hidden state variables like airflow, which cannot be easily measured.
Furthermore, the individual downwash induced by the rotors interacts with both the frame and the blades depending on the current state of the platform. 
The repeatability of tracking errors observed in prior work~\cite{punjani2015deep, ryou2020multi, bansal2016learning} and in this work when performing aggressive maneuvers suggests that the difficulty of learning quadrotor dynamics is not caused by stochasticity in the dynamics, but rather by unobserved state variables such as airflow.

\begin{figure}[t]
    \centering
    \vspace*{6pt}
    \begin{tikzpicture}[>=latex, xscale=1.03]
\tikzstyle{every node}=[font=\footnotesize]

\draw [fill=c3, draw=none](-1.2,2.1) rectangle ++ (3.2,0.5) node [black, anchor=north east, opacity=1] {Our Approach};
\draw [fill=c3, draw=none](2.0,-2.25) rectangle ++ (4.8,4.85);

\draw [fill=c1, draw=none](2.15,0.1) rectangle ++ (4.5,2.4) node [black, anchor=north east, opacity=1] {First Principles};

\draw [fill=c2, draw=none](2.15,-0.1) rectangle ++ (4.5,-2.0) node [black, anchor=south east, opacity=1] {Learning Based};

\node [minimum width=1.5cm, minimum height=1.5cm, align=center, text width = 1.2cm, draw, text depth = 1.0cm, inner sep=0pt, outer sep=0pt] (mm) at (3, 1.1) { MM 
\vspace*{3mm}

\begin{tikzpicture}[>=stealth, yscale=0.6]
\draw [->] (0,0) -- ++ (1,0);
\draw [->] (0,0) -- ++ (0,1);
\draw[smooth, domain=0:0.9] plot (\x,{0.9*(1-exp(-5*\x))});
\end{tikzpicture}
} ;
\node [minimum width=1.5cm, minimum height=1.5cm, align=center, text width = 1.2cm, draw, text depth = 1.0cm, inner sep=0pt, outer sep=0pt] (bm) at (5, 1.1) { 
RM 
\vspace*{3mm}

\begin{tikzpicture}[>=stealth, yscale=0.75]
\begin{scope}
\draw [fill] (0,0) circle (0.025);
\draw [->, decorate,decoration={snake,amplitude=.1mm,segment length=2mm}] (-0.5, -0.1) -- ++ (0, -0.8);
\draw [->, decorate,decoration={snake,amplitude=.1mm,segment length=2mm}] (-0.375, -0.1) -- ++ (0, -0.65);
\draw [->, decorate,decoration={snake,amplitude=.1mm,segment length=2mm}] (-0.25, -0.1) -- ++ (0, -0.55);
\draw [->, decorate,decoration={snake,amplitude=.1mm,segment length=2mm}] (-0.125, -0.1) -- ++ (0, -0.5);
\draw [->, decorate,decoration={snake,amplitude=.1mm,segment length=2mm}] (0, -0.1) -- ++ (0, -0.47);
\draw [->, decorate,decoration={snake,amplitude=.1mm,segment length=2mm}] (0.125, -0.1) -- ++ (0, -0.5);
\draw [->, decorate,decoration={snake,amplitude=.1mm,segment length=2mm}] (0.25, -0.1) -- ++ (0, -0.55);
\draw [->, decorate,decoration={snake,amplitude=.1mm,segment length=2mm}] (0.375, -0.1) -- ++ (0, -0.65);
\draw [->, decorate,decoration={snake,amplitude=.1mm,segment length=2mm}] (0.5, -0.1) -- ++ (0, -0.8);

\begin{scope}[xscale=0.6, yscale=0.15]
\draw[domain=0:1, samples=100, variable=\x] plot({\x},{-0.025+(1/2)*sqrt(-2-4*\x^2+2*sqrt(8*\x^2+1))});
\draw[domain=0:1, samples=100, variable=\x] plot({\x},{0.025-(1/2)*sqrt(-2-4*\x^2+2*sqrt(8*\x^2+1))});
\draw[domain=0:1, samples=100, variable=\x]plot({-\x},{-0.025+(1/2)*sqrt(-2-4*\x^2+2*sqrt(8*\x^2+1))});
\draw[domain=0:1, samples=100, variable=\x] plot({-\x},{0.025-(1/2)*sqrt(-2-4*\x^2+2*sqrt(8*\x^2+1))});
\end{scope}
\end{scope};
\end{tikzpicture}} ;
\node [minimum width=2cm, minimum height=1.8cm, align=left, text width = 1.7cm, text depth = 1cm, inner sep = 0pt, outer sep = 0pt] (nn) at (4, -1.1) { NN 

\vspace*{2mm}

\tikzset{%
  every neuron/.style={
    circle,
    draw,
    minimum size=0.1cm,
    inner sep=0pt,
  },
}

\begin{tikzpicture}[x=1.5cm, y=1.5cm, >=stealth, yscale=0.1, xscale=0.14]

\foreach \m/\l [count=\y] in {1,...,6}
  \node [every neuron/.try, neuron \m/.try] (input-\m) at (0,3.5-\y) {};

\foreach \m [count=\y] in {1,...,4}
  \node [every neuron/.try, neuron \m/.try ] (hidden1-\m) at (3,2.5-\y) {};

\foreach \m [count=\y] in {1,2}
  \node [every neuron/.try, neuron \m/.try ] (output-\m) at (6,1.5-\y) {};

\foreach \l [count=\i] in {1,...,6}
  \draw [] (input-\i) -- ++(-1,0);

\foreach \i in {1,...,6}
  \foreach \j in {1,...,4}
    \draw [] (input-\i) -- (hidden1-\j);

\foreach \i in {1,...,4}
  \foreach \j in {1,2}
    \draw [] (hidden1-\i) -- (output-\j);

\foreach \l [count=\i] in {1,2}
  \draw [] (output-\i) -- ++(1,0);

\end{tikzpicture}

};
\draw (nn.north west) -- ([yshift=-0.4cm]nn.north east) -- ([yshift=0.4cm]nn.south east) -- (nn.south west) -- cycle ;

\node [draw, circle, inner sep=0, minimum size = 2mm] (add) at (6.5,0) {$+$};

\draw [->] (mm.east) -- (bm.west);
\draw [->] (bm.east) -| node [near start, align=left] {$\bm f_\text{prop}$\\$\bm\tau_\text{prop}$} (add);
\draw [->] (nn.east) -| node [near start, align=left] {$\bm f_\text{res}$\\$\bm\tau_\text{res}$} (add);
\draw [->] (add) -- node [midway, align=left] {$\bm f$\\$\bm\tau$} ++ (0.8,0);

\draw [fill=c4, fill opacity=0.5, draw=black] (-1.2, 1.8)  -- ++ (1.7, 0) to [out=0, in=180] (mm.west -| 1.5, 10) -- (mm.west) -- (mm.west -| 1.5, 10)  to [out=180, in=0] (0.9, .75) -- ++(-2.1, 0) -- cycle ;
\draw ([xshift=-2mm]mm.west) -- (mm.west);

\draw [fill=c4, fill opacity=0.5, draw=black] (-1.2, 1.2)  -- ++ (1.95, 0) to [out=0, in=180] ([xshift=-11mm]nn.west) -- (nn.west) -- ([xshift=-11mm]nn.west)  to [out=180, in=0] (0.7, -1.8) -- ++(-1.9, 0) -- cycle ;
\draw ([xshift=-5mm]nn.west) -- (nn.west);

\node [align=center] at (0,0) {
$\begin{bmatrix}
\bm\Omega_{k, \text{cmd}}
\end{bmatrix}$ \\[6pt]
$\begin{bmatrix}
\bm x_k & \bm \Omega_k\\[3pt]
\bm x_{k-1} & \bm \Omega_{k-1} \\[3pt]
\bm x_{k-2} & \bm \Omega_{k-2} \\[3pt]
\vdots & \vdots \\[-4.5pt]
\vdots & \vdots \\[3pt]
\bm x_{k-h} & \bm \Omega_{k-h} 
\end{bmatrix}$};

\end{tikzpicture}
    \vspace*{-6pt}
    \caption{Overview of the proposed architecture to predict aerodynamic forces and torques. The physical modeling pipeline (upper part) consists of a motor model (MM) and a rotor model (RM)---detailed in Sections~\ref{sec:quadratic_model}~and~\ref{sec:bem_model}. It takes the current state $\bm x_k$, current motor speeds $\bm\Omega_k$, and the motor speed command $\bm\Omega_{k,\text{cmd}}$ as an input. Combined with the estimate of the residual forces and torques predicted by the neural network (NN) using the current and past $h$ states, the acting force $\bm f$ and torque $\bm \tau$ are calculated.}
    \vspace*{-18pt}
    \label{fig:architecture}
\end{figure}
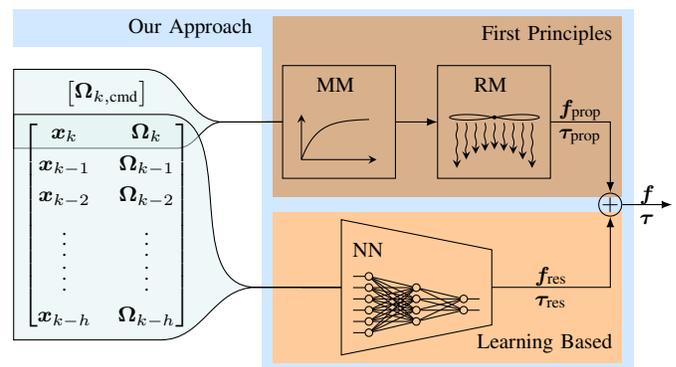

Traditional approaches to quadrotor modeling limit the captured effects to simple linear drag approximations and quadratic thrust curves~\cite{furrer2016rotors, shah2018airsim, faessler2017differential}.
Such approximations are computationally efficient and describe the platform well in low-speed regimes, but exhibit increasing bias at higher velocities as they neglect the influence of the inflow velocity on the generated thrust. 
More elaborate models based on blade-element-momentum (BEM) theory manage to accurately model single rotors at high wind velocities, but they do not account for the aerodynamic interactions between rotors and the frame. 
Parametric gray-box models~\cite{sun2019quadrotor} aim to overcome these limitations by describing the forces and torques as a linear combination of library functions. 
While these models can perform well, their performance hinges on the appropriate choice of basis functions, which require human expert knowledge to design.
Recent research has investigated computational fluid dynamics~\cite{ventura2018high} to model the aerodynamic effects at play during different flight conditions.
While being very accurate, such approaches are computationally expensive and need hours of processing on a compute cluster, rendering them impractical for experiments spanning more than a few seconds. 

Accurately predicting forces acting on the quadrotor at high speeds requires to implicitly estimate the airflow around the vehicle. 
Although this state variable cannot be directly observed, it can be deduced from a sequence of measurements of other observable state variables. 
Thus, learning a high-order dynamics model requires a method for regression of a nonlinear function in a high-dimensional input space. 
Deep neural networks have shown to excel at such high-dimensional regression tasks and have already been applied to dynamic system modeling~\cite{punjani2015deep, bansal2016learning, mohajerin2018deep, portwood2019turbulence, grzeszczuk1998neuroanimator}.
Despite showing promising performance, such purely-learned models require large amounts of data and require careful regularization to avoid overfitting.

\vspace*{-2pt}
\subsection*{Contribution}
\vspace*{-2pt}
This work proposes a quadrotor dynamics model that can accurately capture complex aerodynamic effects by \textit{combining} a state-of-the-art rotor model based on BEM theory with learned residual force and torque terms represented by a deep neural network.
The resulting hybrid model benefits from the expressive power of deep neural networks and the generalizability of first-principles modeling.
The  latter reduces the need for extreme amounts of training data.
The model is identified using data collected from a large set of maneuvers performed on a real quadrotor platform. 
Leveraging one of the biggest optical tracking volumes in the world, the platform's state as well as the motor speeds are recorded during flight. 
The resulting dataset contains 96 flights with a cumulative time of 1h 15min and 1.8 million data points, covering the entire performance envelope of the platform up to observed speeds of $\SI{65}{\kilo\meter\per\hour}$ ($\SI{18}{\meter\per\second}$) and accelerations of $\SI{46.8}{\meter\per\second\squared}$.

The proposed model is compared against state-of-the-art modeling approaches on unseen test maneuvers.
The comparison is done in terms of both evaluation of predicted aerodynamic forces and torques and closed-loop integration of the model in a simulator, each evaluated against real-world reference data. In both categories, a performance increase by a factor of two is observed.

\vspace*{-2pt}
\section{Related Work}\label{sec:related_work}
\vspace*{-2pt}
Traditionally, a rotor is assumed to produce thrust and axial torque proportional to the square of its angular rate with a constant coefficient~\cite{mahony2012multirotor}, which is referred to hereinafter as the simple quadratic model.
While these assumptions are valid for rotors on a static thrust stand and for near-hover flight, they do neither account for the case where the rotors move through air, nor for rotor-to-rotor and rotor-to-body interactions.
Nevertheless, due to its simplicity, this model is still used in well-known aerial robotics simulators such as AirSim~\cite{shah2018airsim}, Flightmare~\cite{song2020flightmare}, RotorS~\cite{furrer2016rotors} and others~\cite{meyer2012comprehensive}.
To improve the accuracy of the thrust model in non-stationary flights over the simple quadratic model, momentum theory has been used in
\cite{hoffmann2007quadrotor,huang2009aerodynamics,hoffmann2011precision}.
Blade element theory is another approach to model a single rotor more accurately. The forces and torques acting on each infinitesimal portion of the blade are integrated over the whole propeller \cite{prouty1995helicopter}. 
This theory has been adopted to model aerodynamic effects on a quadrotor in many studies \cite{orsag2012influence,bristeau2009role,kaya2014aerodynamic,tang2015dynamic,powers2013influence}. 
However, both blade-element theory and momentum theory require the value of the induced velocity which is challenging to estimate. Hence, the blade-element-momentum (BEM) theory is proposed, which combines the above two theories to alleviate the difficulty of calculating the induced velocity.
The resulting model can accurately capture aerodynamic forces and torques acting on single rotors in a wide range of operating conditions~\cite{khan2013toward,gill2017propeller, gill2019computationally}.

Even though BEM outperforms simple quadratic models and often achieves accurate predictions, it does not account for any interaction between the flow tubes of different propellers or the frame~\cite{sun2019quadrotor}.
Previous work has incorporated interaction effects using either static wind tunnel tests~\cite{russell2016wind,schiano2014towards,baris2019wind} where the vehicle is rigidly mounted on a force sensor, or by performing fast maneuvers in instrumented tracking volumes~\cite{torrente2021data}. 
In~\cite{torrente2021data} a simple quadratic model is combined with residual forces predicted by Gaussian Processes. 
While this approach offers a lightweight solution to learn residual forces and can be used for control, it does not model residual torques, effectively neglecting moments caused by rotor-to-rotor interactions.
In~\cite{sun2019quadrotor}, the quadrotor platform is identified using a gray-box model that uses a library of polynomials as basis functions and is able to model both aerodynamic forces and torques. 
This method relies on the predefined function library and also contains discontinuities in the learned model.
Another line of works investigates the modeling of quadrotors using computational fluid dynamics~(CFD)~\cite{ventura2018high, luo2015novel}.
While such simulations achieve results that are highly accurate and manage to capture real-world effects well, they require large amounts of computation time on high-performance compute clusters.

Due to their ability to identify patterns in large amounts of data, deep neural networks represent a promising approach to model aerodynamic effects precisely and computationally efficient. 
A recent line of works employs deep neural networks to learn quadrotor dynamics model purely from data, for both continuous time formulations~\cite{bansal2016learning, punjani2015deep} as well as discrete-time formulations~\cite{mohajerin2018deep, mohajerin2019multistep,shi2019learnedlanding,portwood2019turbulence}.
While approaches relying entirely on learning-based methods have high representative power and the potential to also learn complex interaction effects, they require large amounts of data to train and careful regularization to avoid overfitting. 
    
The approach presented in this work is inspired by~\cite{punjani2015deep, bansal2016learning}, but instead of learning the full dynamics, it combines state-of-the-art BEM modeling based on first principles with a data-driven approach to learn the residual force and torque terms. 
The resulting model benefits from the strong generalization performance of traditional first-principle modeling and the flexibility of learning-based function approximation.

\vspace*{-2pt}
\section{Quadrotor Model}
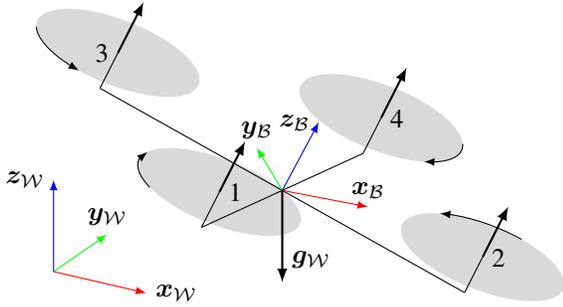
\begin{figure}[t]
\centering
\tdplotsetmaincoords{65}{30}
\begin{tikzpicture}[xscale=0.95,yscale=0.9,tdplot_main_coords, >=latex]
\tikzset{RPY/.style={x={(\nxx,\nxy)},y={(\nyx,\nyy)},z={(\nzx,\nzy)}}}
\def\propz{0.7}
\def\propx{3.3}
\def\propy{2.7}
\def\r{1.3}
\def\l{1.5}
\def\f{1}

\rotateRPY{-10}{20}{0}
\begin{scope}[RPY]

\draw[] (-\propx,0,0) -- (\propx,0,0);
\draw[] (0,-\propy,0) -- (0,\propy,0);

\draw[draw=none, fill=black, opacity=0.15] (0,\propy,\propz) circle (\r);
\draw[draw=none, fill=black, opacity=0.15] (\propx,0,\propz) circle (\r) ;
\draw[draw=none, fill=black, opacity=0.15] (0,-\propy,\propz) circle (\r) ;
\draw[draw=none, fill=black, opacity=0.15] (-\propx,0,\propz) circle (\r) ;
\draw (0,\propy,0) --++ (0,0,\propz) node[right] {4};
\draw (\propx,0,0) --++ (0,0,\propz) node[right] {2};
\draw (0,-\propy,0) --++ (0,0,\propz) node[right] {1};
\draw (-\propx,0,0) --++ (0,0,\propz) node[left] {3};

\draw[thick,->] (0,\propy,\propz) -- ++(0,0,\f);
\draw[thick,->] (\propx,0,\propz) -- ++(0,0,\f);
\draw[thick,->] (0,-\propy,\propz) -- ++(0,0,\f);
\draw[thick,->] (-\propx,0,\propz) -- ++(0,0,\f);	

\def\psi{60}
\def\xa{30}
\def\xb{90}
\def\xc{-120}
\def\xd{210}
\path (\r,\propy,\propz) arc (0:\xa:\r) coordinate (a);
\draw [->] (a) arc (\xa:\xa-\psi:\r) [draw=black] ;

\path (\propx,\r,\propz) arc (90:\xb:\r) coordinate (b);
\draw [->] (b) arc (\xb:\xb+\psi:\r) [draw=black] ;

\path (\r,-\propy,\propz) arc (0:\xc:\r) coordinate (c);
\draw [->] (c) arc (\xc:\xc-\psi:\r) [draw=black] ;

\path (-\propx,\r,\propz)  arc (90:\xd:\r) coordinate (d);
\draw [->] (d) arc (\xd:\xd+\psi:\r) [draw=black] ;

\rotateRPY{0}{0}{45}
\begin{scope}[RPY]
\draw[->,color=red,text=black] (0,0,0) -- ++ (\l,0,0) node[above] {$\bm x_\bfr$};
\draw[->,color=green,text=black] (0,0,0) -- ++ (0,\l,0) node[above] {$\bm y_\bfr$};	
\draw[->,color=blue,text=black] (0,0,0) -- ++ (0,0,\l) node[left] {$\bm z_\bfr$};		
\end{scope}
\end{scope}

\begin{scope}[xshift=-3.2cm, yshift=-1.2cm]
\draw[->,color=red,text=black] (0,0,0) -- ++ (\l,0,0) node[right] {$\bm x_\wfr$};
\draw[->,color=green,text=black] (0,0,0) -- ++ (0,\l,0) node[above] {$\bm y_\wfr$};	
\draw[->,color=blue,text=black] (0,0,0) -- ++ (0,0,\l) node[left] {$\bm z_\wfr$};		
\end{scope}

\draw[thick,->,color=black,text=black] (0,0,0) -- node[right, near end] {$\bm g_\wfr$} ++ (0,0,-1.5*\f);

\end{tikzpicture}
\caption{Diagram of the quadrotor model depicting the world and body frames and illustrating the propeller numbering convention.}
\vspace*{-18pt}
\label{fig:quad_schematic}
\end{figure}

\vspace*{-2pt}
This section explains the hybrid quadrotor model proposed in this work. 
It starts by introducing the notation and the rigid body dynamics (Figure \ref{fig:quad_schematic}), proceeds to explaining two approaches to single-rotor modeling of increasing complexity and concludes with the learned residual model.
The hybrid structure of the model, illustrated in Figure~\ref{fig:architecture}, consists of a \textit{rotor model} and a learned correction. 

\vspace*{-2pt}
\subsection{Notation}
\vspace*{-2pt}
Scalars are denoted in non-bold~$[s, S]$, vectors in lowercase bold~$\bm{v}$, and matrices in uppercase bold~$\bm{M}$.
World $\wfr$ and Body $\bfr$ frames are defined with orthonormal basis i.e. $\{\bm{x}_\wfr, \bm{y}_\wfr, \bm{z}_\wfr\}$.
The frame $\bfr$ is located at the center of mass of the quadrotor.
A vector from coordinate $\bm{p}_1$ to $\bm{p}_2$ expressed in the $\wfr$ frame is written as: ${}_\wfr\bm{v}_{12}$.
If the vector's origin coincides with the frame it is described in, the frame index is dropped, e.g. the quadrotor position is denoted as $\bm{p}_{\wfr\bfr}$.
Furthermore, unit quaternions $\bm{q} = (q_w, q_x, q_y, q_z)$ with $\|\bm{q}\| = 1$ are used to represent orientations, such as the attitude state of the quadrotor body $\bm{q}_{\wfr\bfr}$.
Finally, full SE3 transformations, such as changing the frame of reference from body to world for a point $\bm{p}_{B1}$, can be described by $_\wfr\bm{p}_{B1} = {}_\wfr\bm{t}_{\wfr\bfr} + \bm{q}_{\wfr\bfr} \odot \bm{p}_{B1}$.
Note the quaternion-vector product is denoted by $\odot$ representing a rotation of the vector by the quaternion as in $\bm{q} \odot \bm{v} = \bm{q} \bm{v} \bar{\bm{q}}$, where $\bar{\bm{q}}$ is the quaternion's conjugate.

\vspace*{-2pt}
\subsection{Quadrotor Dynamics}
\vspace*{-2pt}
The quadrotor is assumed to be a 6 degree-of-freedom rigid body of mass $m$ and diagonal moment of inertia matrix $\bm{J}=\mathrm{diag}(J_x, J_y, J_z)$.
The state space is thus 13-dimensional and its dynamics can be written as:
\vspace*{-6pt}
\begin{align}
\small
\label{eq:3d_quad_dynamics}
\dot{\bm{x}} =
\begin{bmatrix}
\dot{\bm{p}}_{\wfr\bfr} \\  
\dot{\bm{q}}_{\wfr\bfr} \\
\dot{\bm{v}}_{\wfr\bfr} \\
\dot{\boldsymbol\omega}_\bfr
\end{bmatrix} = 
\begin{bmatrix}
\bm{v}_\wfr \\  
\bm{q}_{\wfr\bfr} \cdot \mat{0 \\ \bm{\omega}_\bfr/2} \\
\frac{1}{m} \Big(\bm{q}_{\wfr\bfr} \odot \big( \underbrace{\bm{f}_\text{prop} + \bm{f}_\text{res}}_{:= \bm f}\big)\Big) +\bm{g}_\wfr  \\
\bm{J}^{-1}\big( \underbrace{\boldsymbol{\tau}_\text{prop} + \boldsymbol{\tau}_\text{res}}_{:=\boldsymbol \tau}- \boldsymbol\omega_\bfr \times \bm{J}\boldsymbol\omega_\bfr \big)
\end{bmatrix} \; ,
\end{align}
where $\bm{g}_\wfr= [0, 0, \SI{-9.81}{\meter\per\second^2}]^\intercal$ denotes earth's gravity, $\bm{f}_\text{prop}$ is the collective force produced by the propellers including any parasitic effects the rotor model can simulate (e.g. induced drag), and $\bm{f}_\text{res}$ denotes residual forces that are not explained by the rotor model used. 
Similarly, $\boldsymbol{\tau}_\text{prop}$ and $\boldsymbol{\tau}_\text{res}$ are the cumulative torques acting on the platform due to the propellers and residual torques that are not explained by the rotor model. 
\begin{align}
   \bm{f}_\text{prop} &= \sum_i \bm{f}_i \\
   \boldsymbol{\tau}_\text{prop} &= \sum_i \boldsymbol{\tau}_i + \bm{r}_{\text{P},i} \times \bm{f}_i \; ,
\end{align}
\vspace*{-8pt}

\noindent where $\bm{r}_{\text{P},i}$ is the location of propeller $i$ expressed in the body frame and $\bm{f}_i$, $\boldsymbol{\tau}_i$ are the forces and torques generated by the $i$-th propeller.
The rotor models aim at predicting accurate estimates of the single-rotor forces and torques $\bm{f}_i$, $\boldsymbol{\tau}_{i}$, as explained in the following sections. 
The force and torque effects of the fuselage, body, and rotor interaction are not explicitly modeled, but should be captured by the residual dynamics $\bm{f}_\text{res}$ and $\boldsymbol{\tau}_\text{res}$, predicted by a neural network.

\subsection{Rotor Model: Quadratic}\label{sec:quadratic_model}
The simplest model for single propeller is a quadratic fit which assumes the thrust and torque produced by a single propeller to be proportional to the square of its rotational rate (propeller speed) $\Omega$. 
\begin{equation}
\begin{aligned}
    \bm{f}_i(\Omega) = \mat{0 \\ 0 \\ c_{\text{l,q}}\cdot\Omega^2} && \bm{\tau}_i(\Omega) = \mat{0 \\ 0 \\ c_{\text{d,q}}\cdot\Omega^2}
    \end{aligned}
\end{equation}
The coefficients $c_{\text{l,q}}$ and $c_{\text{d,q}}$ are typically identified using a static propeller test stand.
This simplified model is a good approximation for near-hover flight at near-zero velocity without ceiling or ground effects~\cite{powers2013influence} and explains static thrust-test stand measurements very well.
However, it ignores that ego-motion impacts the lift generated by the propeller.
The induced drag, which depends on the propeller speed and body-relative air velocity, is neglected as well, albeit being the dominant source of drag for quadrotors. 
This is sometimes mitigated by combining the model with a linear drag term such as in~\cite{furrer2016rotors, faessler2017differential}. 

\subsection{Rotor Model: BEM}\label{sec:bem_model}
Compared to the quadratic model, Blade-Element-Momentum-Theory (BEM) accounts for the effects of varying relative air speed on the rotor thrust. 
It assumes interaction effects between individual rotors to be negligible and describes each rotor separately. 
The approach presented here is based on classical propeller modeling for helicopters~\cite{prouty1995helicopter}.
The modeling of the propeller lift and drag coefficients is based on~\cite{gill2017propeller, ducard2014modeling}.

First, basic momentum theory is introduced and used to relate the thrust force to a given velocity difference in a flow-tube across the rotor.
Then the aerodynamic blade-element model is presented.
While momentum theory uses the momentum conservation to relate induced airspeed and generated thrust, a blade-element model sums the contributions of infinitesimal blade elements to the total thrust force and drag torque. 
Finally, the full algorithm combining momentum theory and the blade-element model is presented.

For now, the induced velocity $v_i$ is considered to be known as momentum theory and the blade element model jointly yield this information. Together with the known ego-motion of the quadcopter, this fully determines the wind field around the individual propellers.

\mypara{Momentum Theory}
The most simple theory for analyzing rotors is momentum theory as it allows to calculate the thrust of a propeller based on a momentum balance across the rotor.
This balance is done inside a flow-tube with radius~$R$ that fully contains the propeller. 
Assuming a known and constant induced velocity $v_i$ across the diameter of the flow tube, the thrust $T$ of a rotor is given by~\cite{prouty1995helicopter}:
\begin{equation}
T = 2 v_i \rho A \sqrt{v_\text{hor}^2 + (v_\text{ver} - v_i)^2} \; ,
\label{eq:Tmom}
\end{equation}
where $v_\text{hor}$ and $v_\text{ver}$ denote the horizontal and vertical velocity component of the flow tube. 
Note that momentum theory alone does not provide any means for calculating the induced velocity, it merely relates the induced velocity and the thrust based on a momentum balance and does not make any assumption on the physical process that actually accelerates the air.

\begin{figure}[t]
\centerline{
\begin{tikzpicture}[>=stealth, font=\footnotesize, x=(135:0.5cm), y=(0:1cm), z=(90:1cm)]
\tikzset{RPY/.style={x={(\nxx,\nxy)},y={(\nyx,\nyy)},z={(\nzx,\nzy)}}}
\def\r{3.5}
\def\c{0.2}
\def\a{-20}
\def\b{-15}
\def\g{-20}
\def\vi{1.5}
\def\vel{3}
\pgfmathsetmacro\ra{\r/2}
\pgfmathsetmacro\cosb{cos(\b)}
\pgfmathsetmacro\sinb{sin(\b)}
\pgfmathsetmacro\cosg{cos(\g)}
\pgfmathsetmacro\sing{sin(\g)}

\draw [fill=white!90!black, fill opacity = 0.25] (0,0,0) circle (\r);

\rotateRPY{\b}{\a}{0}
\begin{scope}[RPY]
\draw [fill=white!90!black, fill opacity = 0.5] (0,0,0) circle (\r);
\draw [fill=black] (0,0,0) -- (0,-\r,0) -- (\c,-\r,0) -- (\c,0,0) -- cycle;
\draw (0,0,0) -- (\r, 0,0);
\draw (0,0,0) -- (0,\r,0);
\end{scope}
\begin{scope}[canvas is xy plane at z=0]
\draw (0,0) -- (\r, 0);
\draw (0,0) -- (0,\r);
\draw [dashed, -latex]  (-0.5*\r, 0) -- (1.3*\r,0) node [above] {$x_\pfr$};
\draw [dashed, -latex]  (0, -0.93*\r) -- (0, 1.2*\r) node [above] {$y_\pfr$};
\draw [->]  (0,0) -- (-0.9*\r,0) node [above right=-1mm] {$v_\text{hor}$};
\draw [->] (-0.4*\r,0) arc(180:270:0.4*\r) node [pos=0.5, below left =-1mm] {$\Psi$};
\end{scope}

\begin{scope}[canvas is xz plane at y=0]
\draw [->] (\ra,0) arc(0:-\a:\ra)  node [midway, above left] {$a_1$};
\draw [dashed, -latex]  (0, 0) -- (0, -0.75*\r) node [below] {$z_\pfr$};
\draw [->]  (0, 0) -- (0, 0.4*\r) node [right] {$v_\text{ver}$};
\draw [->] (0,0) -- (0, -\vi) node [left] {$v_i$};
\end{scope}
\begin{scope}[canvas is yz plane at x=0]
\draw [->] (\ra,0) arc(0:0.85*\b:\ra) node [midway, right] {$b_1$};
\draw [dashed] (-0.93*\r,0) -- (-0.93*\r,-0.9*\r*\sinb);
\end{scope}
\end{tikzpicture}}
\caption{Lateral flapping $b_1$ and longitudinal flapping $a_1$ occur due to lift imbalance. The azimuth angle $\Psi$ of the blade is measured `from the tail' in the direction of rotation. The horizontal velocity $v_\text{hor}$ and vertical velocity $v_\text{ver}$ are defined opposite to the propeller frame, i.e. if the propeller moves along $z_\pfr$ the relative velocity will also have a positive vertical component. Note that coning is not shown to improve the clarity of the schematic.}
\vspace*{-6pt}
\label{fig:flappingCoordinate}
\end{figure}
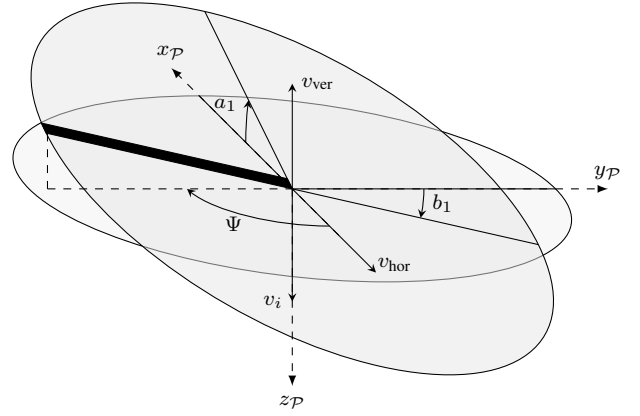

\mypara{Blade Element Theory}
The main purpose of a blade element model is to estimate the acting forces and torques accurately. A propeller consists of $b$ identical blades (typically $b=2$ or $b=3$) attached to the rotor hub, each acting as a wing producing lift and drag forces.
The propeller coordinate frame $\pfr$ shown in Figure~\ref{fig:flappingCoordinate} is defined such that the $z_\pfr$-axis points down and the $x_\pfr$-axis opposes the horizontal component $v_\text{hor}$ of the incoming wind.

The finite stiffness of the blade and its hub-mount allow it to bend and deform, transmitting forces and introducing torques around the hub. It also causes the rotor-disk plane to be tilted with respect to the propeller coordinate system. 
This deformation can be split into a symmetric \emph{coning} component $a_0$ due to the overall lift produced by the propeller (not shown in Figure~\ref{fig:flappingCoordinate}) and an asymmetric \emph{flapping} component that depends on the azimuth angle $\Psi$ of the propeller.
Figure~\ref{fig:flappingCoordinate} illustrates this: in forward flight one side of the propeller experiences a higher relative airspeed (advancing blade) compared to the opposite side (retracting blade).
Thus, there is a lift imbalance between the sides of the propeller which in turn causes the elastic propeller to bend upwards on the advancing side. 
This is called lateral flapping with the associated flapping angle $b_1$. 
Due to the inertia of the blade, the lateral flapping also induces a longitudinal flapping angle $a_1$.

\begin{figure}[t]
\centering
\centerline{
    \begin{tikzpicture}[>=stealth, font=\footnotesize]
\def\thetaval{-15}
\def\phival{20}
\def\lift{2}
\def\drag{1.5}
\def\vel{3}
\def\vi{1}
\pgfmathsetmacro\r{\vel/2}
\pgfmathsetmacro\alphaval{\thetaval + \phival}
\pgfmathsetmacro\cosphi{cos(\phival)}
\pgfmathsetmacro\sinphi{sin(\phival)}
\pgfmathsetmacro\thrust{\lift*\cosphi + \drag*\sinphi)}
\pgfmathsetmacro\hforce{-\lift*\sinphi + \drag*\cosphi)}

\begin{scope}[rotate=\phival]
\draw [->] (-\vel,0) -- node [pos=0.4, below] {$U$} (0,0);
\draw [dashed] (-1.4*\vel,0) -- (0,0);
\draw [->] (-1.3*\vel, 0) arc (180:180-\phival+\thetaval:1.3*\vel) node [midway, right] {$\alpha$};
\draw [->] (0,0) -- node [midway, above] {$\mathrm{d}D$} (\drag,0);
\draw [->] (0,0) -- node [near end, right] {$\mathrm{d}L$} (0, \lift);
\draw (0,0.2) -| (0.2,0);
\end{scope}

\begin{scope}[rotate=\thetaval]
\draw [dashed] (-1.4*\vel,0) -- (0,0);

\begin{scope}[scale=1.35*\vel, xshift=-0.33cm]
\draw (0,0) -- node [pos = 0.2, above = -0.5mm] {$c$} (1,0); 
\draw (0,-0.015) -- ++ (0, 0.03);
\draw (1,-0.015) -- ++ (0, 0.03);
\input{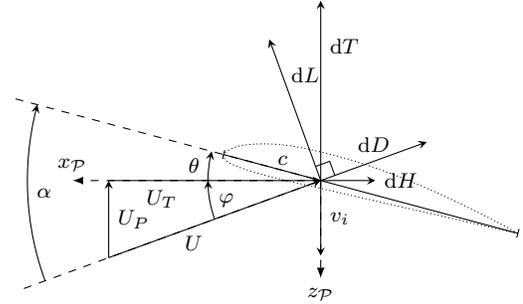}
\end{scope}

\end{scope}

\draw [->] (-\vel*\cosphi,0) -- node [below, near start] {$U_T$} (0,0);
\draw [->] (-\vel*\cosphi,-\vel*\sinphi) -- node [right, midway] {$U_P$}  (-\vel*\cosphi,0);
\draw [<-] (-\r, 0) arc(180:180+\phival:\r) node [right, midway] {$\varphi$};
\draw [->] (-\r, 0) arc(180:180+\thetaval:\r) node [left, midway] {$\theta$};

\draw [->] (0,0) -- node [near end, right] {$\mathrm{d}T$} (0, \thrust);
\draw [->] (0,0) --  (\hforce, 0) node [right] {$\mathrm{d}H$};
\draw [->] (0,0) -- node [midway, right] {$v_i$} (0, -\vi);

\draw [dashed, -latex] (0,0) -- node [pos=1, above] {$x_\pfr$}  (-1.1*\vel,0);
\draw [dashed, -latex] (0,0) -- (0, -1.3*\vi) node [below] {$z_\pfr$};
\end{tikzpicture}
}
\caption{A blade element located at radius $r$ and azimuth angle $\Psi$.}
\vspace*{-12pt}
\label{fig:bladeElement}
\end{figure}

Figure~\ref{fig:bladeElement} shows the velocities, angles and forces of a blade element. 
The chord of the airfoil is rotated by an angle $\theta$ relative to the $xy$-plane. 
Together with the inflow angle $\varphi$, this results in a total angle of attack $\alpha = \theta + \varphi$. 
Each blade element produces a lift force $\mathrm{d}L$ perpendicular to the incoming airstream and a drag force $\mathrm{d}D$ in the direction of the incoming airflow. 
The thrust force $\mathrm{d}T$ and horizontal force $\mathrm{d}H$ are aligned with the propeller coordinate frame.

The tangential velocity $U_T$ and parallel-to-motor velocity $U_P$ are related to the angular velocity $\Omega$ of the propeller as
\begin{align}
U_T(r, \Psi) &= \Omega r + v_\text{hor} \sin \Psi \\
U_P(r, \Psi) &= v_\text{ver} - v_i 
\\\nonumber &\hphantom{=} -r \Omega (a_1 \sin \Psi + b_1 \cos \Psi) 
\\\nonumber &\hphantom{=} +v_\text{ver} (a_0 - a_1 \cos \Psi - b_1 \sin \Psi) \cos \Psi  %
\; .
\end{align}
The local angle of attack $\alpha$ can be calculated as \\[-18pt]
\begin{align}
\varphi(r, \Psi) &= \arctan(U_P(r, \Psi) / U_T(r, \Psi)) \label{eq:bemStart}\\
\alpha(r, \Psi) &= \theta_0 + \frac{r}{R} \theta_1 + \varphi(r, \Psi) \; ,
\end{align}
where $\theta_0$ is the pitch angle of the blade and $\theta_1$ the blade twist.
The differential lift force $dL$ and the differential drag $dD$ can be expressed as functions of radius $r$ and azimuth angle $\Psi$:
\begin{align}
dL(r, \Psi) &= c(r) c_l(\alpha(r, \Psi)) (U_T(r, \Psi)^2 + U_P(r, \Psi)^2)  
\label{eq:diffLift} \\
dD(r, \Psi) &= c(r) c_d(\alpha(r, \Psi)) (U_T(r, \Psi)^2 + U_P(r, \Psi)^2)  ,
\label{eq:diffDrag}
\end{align}
where $c(r)$ is the chord length and $c_l(\alpha)$, $c_d(\alpha)$ are the angle-of-attack dependent coefficients of lift and drag respectively. 
The coefficients are modeled as proposed in \cite{gill2017propeller, ducard2014modeling} as
\begin{align}
    c_d(\alpha) = c_{d,0} \sin^2\alpha && \qquad c_l(\alpha) = c_{l,0} \sin\alpha\cos\alpha \; ,
\end{align}
where $c_{d,0}$ and $c_{l,0}$ are experimentally determined by measuring the lift and drag torque on a thrust test stand.

The overall thrust $T$, horizontal force $H$ and drag torque $Q$ are obtained through integration.
\begin{align}
T &= \frac{b \rho}{4\pi} \int_0^R \int_0^{2\pi} \! dL \cos \phi + dD \sin \phi ~ \mathrm{d}\Psi~ \mathrm{d}r  \label{eq:Tbem} \\
H &= \frac{b \rho}{4\pi} \int_0^R \int_0^{2\pi} \!\!\left(-dL \sin \phi + dD \cos \phi\right)\sin \Psi \;\mathrm{d}\Psi \; \mathrm{d}r \\
Q &= \frac{b \rho}{4\pi} \int_0^R \int_0^{2\pi} \!\! \left(-dL \sin \phi + dD \cos \phi\right) r ~ \mathrm{d}\Psi ~\mathrm{d}r
\label{eq:Qbem}
\end{align}

\begin{figure}[t]
\centerline{
\begin{tikzpicture}[>=stealth]
\def\hubwidth{0.5}
\def\hubheight{1}
\def\propthickness{0.1}
\def\ef{0.3}
\def\r{4}
\def\alpha{-15}
\def\tr{0.2}

\draw [draw=black, fill=gray, fill opacity = 0.5] (-0.5*\hubwidth,0) rectangle ++ (\hubwidth,-\hubheight);
\draw [dash pattern={on 7pt off 2pt on 1pt off 3pt}] (0,-1.2*\hubheight) -- ++ (0, 1.6*\hubheight);
\draw (-\ef*\r,0) rectangle (\ef*\r,\propthickness);
\draw [<->] (0,0.25) -- node [above, midway] {$e$} ++ (-\ef*\r,0);

\begin{scope}[shift={(\ef*\r,0)}, rotate=\alpha]
\draw [fill] (0,0.5*\propthickness) circle (0.5*\propthickness);
\draw (-0.25*\propthickness,0) rectangle ++ (\r,\propthickness);
\draw [-{>[flex=0.75]}] (\tr,0.5*\propthickness) arc (0:315:\tr) node [below] {$M_\text{spring}$};
\end{scope}

\begin{scope}[shift={(-\ef*\r,0)}, rotate=-\alpha]
\draw [fill] (0,0.5*\propthickness) circle (0.5*\propthickness);
\path [clip] (0.5*\propthickness, -\propthickness) rectangle (-0.3*\r, 2*\propthickness);
\draw (0.25*\propthickness,0) rectangle ++ (-\r,\propthickness);
\end{scope}

\end{tikzpicture}
}
\caption{Illustration of the hinged blade model.}
\vspace*{-16pt}
\label{fig:hubmodel}
\end{figure}
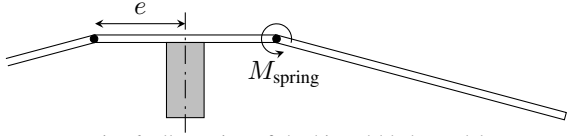

\mypara{Blade Elasticity}
Standard helicopters have their rotor blades connected to the rotor hub through a hinge pin, optionally with an offset. 
This is not true for small rotor sizes typically found on multicopters, since the blade is fixed and elastic which breaks the assumptions made in standard helicopter literature.
Therefore, the model is slightly adapted to capture the characteristics of small propellers: a blade is rigid and connected to the rotor hub with a hinge \emph{and a torsional spring} at an offset $e$ \cite{hoffmann2007quadrotor} as shown in Figure~\ref{fig:hubmodel}. 
The coning angle $a_0$ as well as the flapping angles $a_1$ and $b_1$ can be calculated by equating the moments acting on the rotor hub. 
At the hinge position, the following moment equilibrium occurs: \\[-7pt]
\begin{equation}
0 = M_\text{w} + M_\text{gyro} + M_\text{inertial} + M_\text{cf} + M_\text{aero} + M_\text{spring} \; ,
\label{eq:bem_sumofmoments}
\end{equation}

\vspace*{-3pt}
\noindent where the moment $M_\text{w}$ is caused by the weight of the blade, the moment $M_\text{gyro}$ is due to gyroscopic effects the blade experiences when a non-zero rollrate or pitchrate are present,  $M_\text{inertial}$ comes from the inertia of the blade and its angular acceleration during the flapping motion, $M_\text{cf}$ is caused by centrifugal forces when the blade flaps, the moment $M_\text{aero}$ is a result of the lift generated by the blade, and lastly $M_\text{spring}$ is the restoring moment produced by the hinge spring. 
For brevity, the derivation of the coning and flapping angles are omitted here. They closely follow~\cite{prouty1995helicopter}~(pp. 463). Due to the torsional spring at the hinge, the spring moment needs to be considered additionally: \\[-7pt]
\begin{equation}
M_\text{spring} = k_\beta \left(a_0 + a_1 \cos \Psi + b_1 \sin \Psi\right) \; ,
\end{equation}

\vspace*{-3pt}
\noindent
where $k_\beta$ is the given spring stiffness.
From ~\eqref{eq:bem_sumofmoments}, the coning and flapping angles are calculated. The resulting expression is omitted here for readability.

\mypara{Complete BEM-Model Algorithm}
Throughout above explanations, the induced velocity $v_i$ was treated as a known quantity. 
However, when simulating the vehicle the induced velocity is unknown and needs to be calculated.
This can be done by combining the results from momentum theory and blade-element theory: \eqref{eq:Tmom} and \eqref{eq:Tbem} can both be used to calculate the thrust of the propeller. Due to the relatively stiff blade, the flapping and coning angle are small (typically less than \SI{1}{\degree}) and can thus be neglected as a first approximation to calculate the induced velocity. 

Due to the nature of the physical modeling process, the results from the momentum theory are only valid if the vehicle does not fly in its own downwash. This occurs when the vehicle descends with a certain speed, e.g. the propeller is in vortex-ring state~\cite{hoffmann2007quadrotor} if and only if
\begin{equation}
   0 < \frac{v_{\pfr, z}}{v_i}  < 2  \; .
\end{equation}
If the results from momentum-theory are not applicable, the induced velocity can not be calculated. 
In~\cite{hoffmann2007quadrotor} a solution is presented which relies on an empirical fit to calculate the induced velocity in such flight conditions. 
The proposed fit consists of a quartic polynomial to approximate $v_i$ as follows:
\begin{align}
\tilde{v}_i &= v_{h,i} \left( 1 + 1.125 (v_{\pfr, z}/v_{h,i}) - 1.372 (v_{\pfr, z}/v_{h,i})^2 \right. \label{eq:v1fit} \\\nonumber &\left.+ 1.718 (v_{\pfr, z}/v_{h,i})^3 -0.655 (v_{\pfr, z}/v_{h,i})^4\right) \; ,
\end{align}
where $v_{h,i}$ is the induced velocity if the vehicle would fly horizontally in the given flight state, i.e. set ${v_{\pfr, z} = 0}$. To ensure a smooth transition back to the physical modeling, the final induced velocity in vortex ring state is given as $v_i = \max(\tilde{v}_{i}, v_{h,i})$. \\[2pt]
\noindent The algorithm thus consists of the following steps:
\begin{enumerate}
\item Assume $a_0 = 0, a_1 = 0, b_1 = 0$.
\item Find $v_i$ such that \eqref{eq:Tmom} and \eqref{eq:Tbem} are simultaneously satisfied, i.e. the thrust calculated by momentum theory and blade element theory are identical. If inside vortex-ring state, use the approximation presented above.
\item Calculate the coning and flapping angle $a_0$, $a_1$ and $b_1$ with the previously computed induced velocity.
\item Using the previously calculated induced velocity and blade flapping angles, \eqref{eq:Tbem} -- \eqref{eq:Qbem} can be evaluated again.
\item The total force of the propeller and torque around the center of the propeller are given by:
\begin{align}
\nonumber
\bm f_\pfr = \begin{bmatrix} 
- (H + \sin a_1 T) \\
\pm \sin b_1 T \\
-T \cos a_0
\end{bmatrix} &&
\bm \tau_\pfr = \begin{bmatrix} 
\pm k_\beta b_1 \\
k_\beta a_1 \\
\mp Q
\end{bmatrix} \; ,
\label{eq:bemFinal}
\end{align}
where the upper sign of $\pm$ and $\mp$ corresponds to propeller rotating clockwise and the lower sign needs to be used for a counter-clockwise spinning propeller. 
\end{enumerate}

\subsection{Learned Residual Dynamics}
Both rotor models presented in Sections~\ref{sec:quadratic_model}~and~\ref{sec:bem_model} do not account for aerodynamic forces and torques caused by the quadrotor body or interaction effects between the propellers. 
In this work, these residual dynamics are approximated by a deep neural network. 
Modeling such effects accurately requires to implicitly estimate the airflow around the vehicle. 
Considering the airflow as hidden state of the system, it can be estimated by measuring a history of observable state variables. 
In this work, the angular and linear velocities, as well as motor speeds are used as input features for the neural network. 
A history length of $h = 20$, with temporary equally-spaced samples with a $\delta t = \SI{2.5}{\milli\second}$ is used, effectively giving information of the platform evolution over the past $\SI{50}{\milli\second}$.

\rev{
The network architecture is empirically validated by minimizing the prediction error on an unseen test set. The candidate architectures consist of temporal-convolutional (TCN) encoders~\cite{oord2016wavenet} and fully-connected (MLP) encoders, which both are combined with two fully-connected heads, one for  the residual force prediction and one for the residual torque prediction. 
Each architecture uses leaky-ReLU activations and a linear output layer. 
Training is performed in a supervised fashion using the Adam optimizer by minimizing the RMSE loss on forces and torques between predictions and labels. 
Table~\ref{tab:ablation_study} shows the main results of these ablation experiments. 
Due to its favourable performance versus inference time trade off, the medium-sized temporal-convolutional encoder (TCN-medium) was selected for all subsequent experiments. 
}

\begin{table}
\vspace*{4.5pt}
\caption{\textnormal{Comparison of different network architectures with respect to RMSE of force and torque prediction on a held-out test set.}\\[-4pt]}
\label{tab:ablation_study}
\begin{tabularx}{1\linewidth}{X|ccc}
Architecture & Force RMSE [N] & Torque RMSE [Nm] & \# Param \\
\midrule
TCN small      &0.365 &  6.525 $\times 10^{-3}$ &  12k \\
TCN medium     &0.352 &  5.274 $\times 10^{-3}$ &  25k \\
TCN large      &0.355 &  4.674 $\times 10^{-3}$ &  72k \\
MLP            &0.356 &  5.172 $\times 10^{-3}$ &  30k 
\end{tabularx}
\vspace*{-15pt}
\end{table}

\section{Experimental Setup}\label{sec:data_generation}
\subsection{Data Collection}
To train the model and to verify its accuracy, real world measurement data is needed. It is recorded in a flying arena equipped with a motion tracking system with a usable volume of {$\SI{25}{\meter}\times\SI{25}{\meter}\times\SI{8}{\meter}$}.
The Vicon\footnote{\url{https://www.vicon.com/}} motion tracking system allows to record accurate position and attitude measurements at $\SI{400}{\hertz}$.
Additionally, onboard IMU measurements and motor speeds are recorded at \SI{1}{\kilo\hertz} by the low-level flight controller.
This onboard data and the pose measurements need to be synchronized and fused in post processing. 
For this purpose interpolating cubic splines are fitted to the datapoints, which allows fusing the asynchronous measurements from both data sources, and recovers the full dynamic state.
Furthermore, to estimate the unobserved linear velocity and angular acceleration, differentiation of the fitted splines provides less noisy estimates than direct differentiation of the discrete, noisy measurements.
For the means of time synchronization, offset and clock skew are estimated through the correlation quality of the axis-wise angular rate measurement from the IMU with the spline.
Gyroscope measurements are used because they provide better noise characteristics than the accelerometer data. The clock skew was typically observed to be \SI{2.4}{\percent}.
The motor data is smoothed with a finite-impulse-response fourth-order Butterworth low-pass filter with a cutoff frequency corresponding to the time-constant of the motors, identified from the step response of the motors. This ensures that noise is suppressed without attenuating high-frequency motor signals more than \SI{3}{\decibel}.

The resulting dataset contains 1.8 million data points recorded from 96 flights covering 1h:15min of flight time. 
The dataset is split into 70\% training, 20\% validation, and 10\% test set.
Each subsets contains trajectories that cover the full range of speeds and accelerations observed in the full data set.

\subsection{Quadrotor Platform}

The real-world flights are performed with a custom-made quadrotor platform.
It features an Armattan Chameleon $\SI{6}{inch}$ main frame, equipped with Hobbywing XRotor 2306 motors and $\SI{5}{inch}$, three-bladed propellers.
The platform has a total weight of \SI{752}{\gram} and can produce a maximum static thrust of approximately \SI{33}{\newton}, which results in a static thrust-to-weight ratio of $4.5$.
The weight and power of this platform is comparable to the ones used by professional pilots in drone racing competitions.
The platform's main computational unit is an NVIDIA Jetson TX2 accompanied by a ConnectTech Quasar carrier board.
In all real world flights, control commands in the form of collective thrust and bodyrates are computed on a laptop computer and sent via a Laird module to the Jetson TX2. 
The Jetson then forwards these commands to a commercial flight controller running BetaFlight\footnote{\url{https://github.com/betaflight/betaflight}}, which produces single-rotor commands that are fed to a 4-in-1 electronic speed controller.

\subsection{Control System}
Control commands are produced by a control pipeline consisting of two levels: (\rom{1}) a high-level non-linear quadratic MPC controller generating bodyrate and collective thrust commands at $\SI{100}{\hertz}$, and (\rom{2}) a low-level Betaflight controller tracking the desired bodyrate setpoint at $\SI{1}{\kilo\hertz}$.
To ensure repeatability, \emph{BetaFlight} features targeted at human piloted drones (feed-forward terms) are disabled, and the controller is reduced to a PID with equal parameters for simulation and real-world flight.
BetaFlight is run as a software module within the simulation, resembling the real control system.
Both controllers are kept equal in the simulation with respect to the real-world experiments, to guarantee equal performance in both scenarios.

\subsection{Simulator Extension}\label{sec:method}
To compare different models, their resulting simulation accuracy is evaluated with respect to the real-world trajectory.
For this purpose, the closed-loop system is simulated forward in time.
While the rigid-body dynamics are given by \eqref{eq:3d_quad_dynamics} and the aerodynamic force and torque are provided by the model in question, there are two further  components needed for an accurate simulation: integration and motor dynamics.

\mypara{Integration}
The integration is performed by a symplectic Euler scheme with a timestep of $\SI{1}{\milli\second}$ using the rigid body dynamics~\eqref{eq:3d_quad_dynamics}, the modeled linear and angular accelerations of the tested model, and the motor dynamics explained in the following section.
The advantage of the symplectic Euler scheme is its energy conservation property, which is invalidated with other integration schemes, such as the standard Euler methods or the Runge-Kutta family of integrators.

\mypara{Motor Dynamics}
Since the aerodynamic model is based on the angular speed of the propeller, the motors are modeled as a first-order system according to 
\begin{equation}
\frac{\delta }{\delta t} \Omega = \frac{1}{\tau_\Omega} (\Omega_{cmd} - \Omega)
\end{equation}
where $\Omega_{cmd}$ is the commanded propeller speed and $\tau_\Omega$ is the motor time constant.
For the quadrotor platform used throughout the experiments, the time constant was identified to be $\tau_\Omega = \SI{33}{\milli\second}$.

\section{Experiments and Results}\label{sec:experiments}
The evaluation procedure is designed to address the following questions:
(i)~When does the classical approach to quadrotor modeling based on quadratic thrust and torque curves start to break down?
(ii)~How do the forces and torques predicted by a model based on BEM compare with respect to a simple quadratic model?
(iii)~What is the contribution of a learned residual dynamics component? 
The reader is encouraged to watch the attached video to understand the highly dynamic nature of the experiments.

\subsection{Experimental Setup}
Throughout the experiments multiple models are fitted using the training and validation data, and evaluated using the test data. 
The predictive performance of these models is compared in two different settings, covered in the subsequent sections:
in Section~\ref{sec:open_loop_comparison}, the RMSE of predicted forces and torques is compared on unseen flight data;
in Section~\ref{sec:closed_loop_comparison}, different dynamics models are evaluated in conjunction with a known controller to determine the mismatch between simulation and the real world in a closed-loop scenario. 
Both types of comparisons are performed on unseen test trajectories that cover the entire performance envelope of the platform.
Each trajectory has been performed on the real platform in an instrumented tracking volume as explained in Section~\ref{sec:data_generation}.

\subsection{Comparison of Predictive Performance}\label{sec:open_loop_comparison}
In a first set of experiments, the presented models are compared in terms of predicted forces and torques on unseen trajectories.
These trajectories cover the entire performance envelope of the quadrotor,
ranging from slow near-hover trajectories with speeds below $\SI{5}{\meter\per\second}$ to aggressive trajectories at the limit of the platform's capabilities, exceeding speeds of $\SI{18}{\meter\per\second}$ and accelerations up to $\SI{46.8}{\meter\per\second\squared}$.
The models compared in this experiment consist of the quadratic model~(\textit{Fit}), the BEM model~(\textit{BEM}) and a naive model predicting all zeros~(\textit{None}).
Each of these models is augmented with a learned residual correction using a neural network, marked with~\textit{+NN}. 
While the \textit{None} model represents a naive baseline to better understand the magnitude of prediction errors, \textit{None+NN} illustrates the performance of a purely learned model. 
Finally, the approach presented in~\cite{sun2019quadrotor} is compared, denoted as \textit{PolyFit} as it uses automatically selected polynomial basis functions to fit the model.
\begin{table}[t]
\centering
\caption{\textnormal{Comparison of model performance in terms of RMSE on an unseen test set. Approaches marked with an asterisk are trained on a reduced training set to compare generalization performance.\\}}
\label{tab:forces_torques}
\begin{tabularx}{1\linewidth}{l|CCCC|CC}
Model   & $F_\text{xy}$ [N]     & $F_\text{z}$ [N]     & $M_\text{xy}$  [Nm]   & $M_\text{z}$  [Nm]    & $F$ [N]       & $M$ [Nm]  \\
\midrule
None    & 1.549   & 13.618  & 0.036   & 0.006   & 7.964   & 0.029  \\
Fit     & 1.536   & 1.381   & 0.104   & 0.033   & 1.486   & 0.087  \\
\rev{BEM}     & 0.803   & 1.265   & 0.090   & 0.017   & 0.982   & 0.074  \\
PolyFit~\cite{sun2019quadrotor} & 0.453   & 0.832   & 0.027   & 0.008   & 0.606   & 0.022  \\
None+NN & 0.236   & 0.681   & 0.017   & \textbf{0.002}   & 0.438   & 0.014   \\
Fit+NN  & 0.232   & 0.722   & 0.017   & 0.004   & 0.458   & 0.014  \\
BEM+NN (ours) & \textbf{0.204}   & \textbf{0.504}   & \textbf{0.014}   & 0.004   & \textbf{0.335}   & \textbf{0.012}  \\
\midrule
PolyFit*~\cite{sun2019quadrotor}& 1.450   & 6.637   & 2.815   & 0.164   & 4.011   & 2.301 \\
None+NN*& 0.470   & 1.959   & \textbf{0.007}   & \textbf{0.002}   & 1.194   & \textbf{0.006}   \\
Fit+NN* & 0.501   & 1.225   & 0.024   & 0.013   & 0.817   & 0.021  \\
BEM+NN* (ours) & \textbf{0.344}   & \textbf{0.816}   & 0.025   & 0.008   & \textbf{0.549}   & 0.021  \\
\end{tabularx}
\end{table}
\begin{figure}[t]
    \centering
    \input{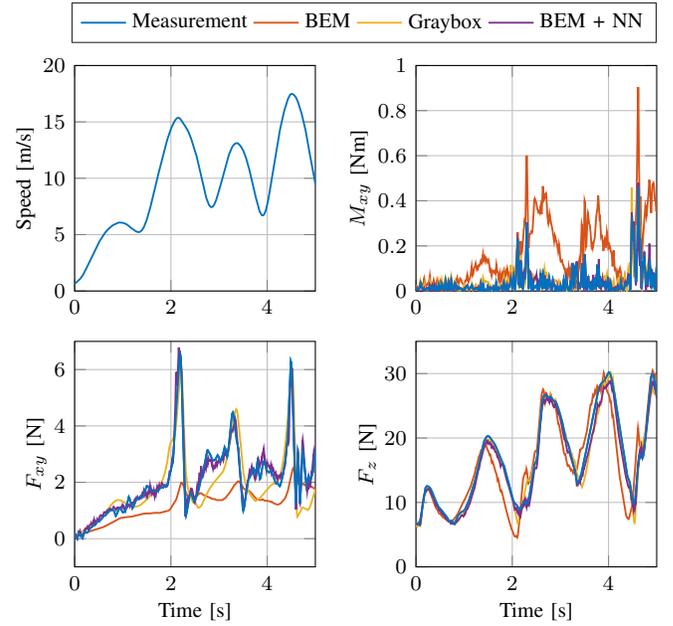}
    \caption{The plot illustrates the results presented in Table~\ref{tab:forces_torques}. The plots show a highly aggressive maneuver (from the test dataset) where only models with neural net augmentation predict the forces and torques well.\vspace*{-12pt}}
    \label{fig:forces}
\end{figure}
\begin{table*}[t]
\caption{\textnormal{Comparison of closed-loop simulation performance on an unseen test set of different trajectories. 
Results show the positional RMSE between trajectories flown in simulation with different dynamics models and the same set of trajectories flown on the real platform.
Models marked with an asterisk~($^\star$) were trained only on slow data up to
\SI{5}{\meter\per\second}.}}
\label{tab:closed_loop_comparison}
\begin{tabularx}{1\linewidth}{l|CC|CCCCCC|CCCC|C}
                   &\rotatebox{90}{$v_\text{mean}$ [m/s]} &\rotatebox{90}{$v_\text{max}$ [m/s]} &\rotatebox{90}{     Fit} &\rotatebox{90}{     BEM} &\rotatebox{90}{PolyFit \cite{sun2019quadrotor}} &\rotatebox{90}{ None+NN} &\rotatebox{90}{  Fit+NN} &\rotatebox{90}{\parbox{1.4cm}{BEM+NN\\ (ours)}} &\rotatebox{90}{PolyFit$^\star$} & \rotatebox{90}{None+NN$^\star$} &\rotatebox{90}{Fit+NN$^\star$} &\rotatebox{90}{\parbox{1.4cm}{BEM+NN$^\star$ \\ (ours)}}
                   &\rotatebox{90}{RotorS \cite{furrer2016rotors}}  \\
\midrule
Lemniscate          &    1.67 &    3.51 &   0.061 &   0.059 &\textbf{   0.043} &   0.046 &   0.049 &   0.059 &  \textbf{0.046} & 0.049 &   0.048 &   0.053 &   0.114  \\
Random Points       &    2.38 &    8.25 &   0.167 &   0.138 &   0.130 &   crash &\textbf{   0.126} &   0.141 &  crash & 0.130 &\textbf{   0.123} &   0.134 &   0.239  \\
Lemniscate         &    3.21 &    7.04 &   0.183 &   0.146 &\textbf{   0.102} &   0.103 &   0.112 &   0.109 &crash & \textbf{   0.100} &   0.112 &   0.112 &   0.148  \\
Melon               &    3.57 &    7.63 &   0.229 &   0.163 &\textbf{   0.117} &   0.126 &   0.126 &   0.133 &crash &\textbf{   0.127} &   0.134 &   0.137 &   0.192  \\
Slanted Circle      &    6.92 &   10.75 &   0.381 &   0.232 &\textbf{   0.166} &   0.172 &   0.168 &   0.167 &  crash & 0.210 &   0.193 &\textbf{   0.185} &   0.257  \\
Linear Oscillation  &    7.25 &   16.95 &   0.506 &   0.438 &   0.172 &   0.270 &   0.234 &\textbf{   0.171} &  crash & 0.258 &   0.263 &\textbf{   0.216} &   0.562  \\
Race Track          &    7.64 &   13.14 &   0.414 &   0.286 &   0.233 &   0.283 &   0.223 &\textbf{   0.214} &  crash & 0.257 &   0.262 &\textbf{   0.240} &   0.424  \\
Melon               &    7.74 &   13.55 &   0.431 &   0.221 &   0.239 &   0.179 &   0.164 &\textbf{   0.155} &  crash & 0.263 &   0.207 &\textbf{   0.185} &   0.398  \\
Slanted Circle      &    8.57 &   13.32 &   0.531 &   0.217 &   0.255 &   0.206 &   0.197 &\textbf{   0.192} &  crash & 0.340 &   0.268 &\textbf{   0.180} &   0.397  \\
Race Track          &    9.94 &   17.81 &   0.617 &   0.408 &   0.820 &   0.447 &   0.320 &\textbf{   0.301} &  crash & 0.446 &   0.420 &\textbf{   0.370} &   0.708  \\
Lemniscate          &   12.01 &   19.83 &   0.762 &   0.549 &   0.316 &   0.782 &   0.352 &\textbf{   0.286} &  crash & 0.469 &   0.423 &\textbf{   0.371} &   1.158  \\
Ellipse             &   15.02 &   19.20 &   0.855 &   0.369 &   crash &   0.347 &   0.402 &\textbf{   0.285} &  crash & 0.605 &   0.481 &\textbf{   0.290} &   0.653  \\
\end{tabularx}
\end{table*}
\begin{figure*}
\centering
\input{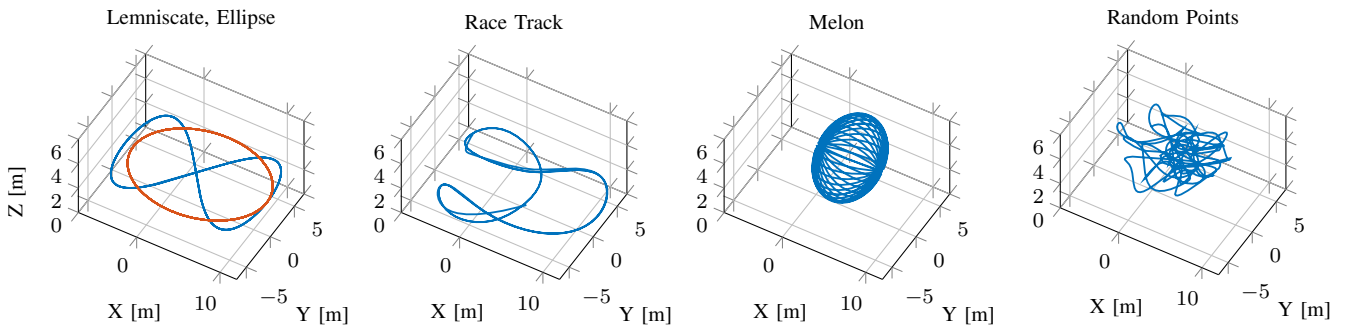}
\vspace*{6pt}
\caption{Trajectories used for testing. Each trajectory is flown multiple times with varying speeds.}
\vspace*{-12pt}
\label{fig:trajectories}
\end{figure*}

To evaluate generalization performance, each approach is trained on two datasets: the entire training set as explained in Section~\ref{sec:data_generation} and a reduced dataset that only covers linear speeds up to \SI{5}{\meter\per\second}. 
This reduced dataset is used to identify new parameters for each approach, which are marked with an asterisk. 

Table~\ref{tab:forces_torques} summarizes the results of this experiment, while Figure~\ref{fig:forces} illustrates performance on a highly aggressive maneuver. 
The proposed hybrid model based on BEM and learned residual dynamics consistently outperforms all other models on the predicted forces. 
Note that the trajectories performed in this work are designed to minimize yaw rate, and as a result only cover extremely small yaw torques~$M_\text{z}$ (the largest yaw torque in the dataset is \SI{0.072}{\newton\meter}), as discussed in Section \ref{sec:discussion}. 
It is evident that thrust and torque along the body $z_\bfr$-axis are more challenging to predict accurately.
The reason for this is two-fold:
First, all linear acceleration actuation lies in the $z_\bfr$ direction, contributing actuation noise predominantly along this axis.
Second, all trajectories are designed to minimize yaw torque, since this is the least actuated torque direction, and significantly limits the acceleration envelope, and therefore the attainable agility.
This explains the good performance of even very naive baselines with respect to this metric. 

When comparing on the reduced dataset (Table~\ref{tab:forces_torques}, bottom), the performance of the proposed approach gracefully degrades, still outperforming the baselines in terms of predicted forces. 
Note that the \textit{PolyFit} baseline completely breaks down in this setting, indicating poor generalization to unseen data. 
The purely learning-based baseline outperforms all other approaches in the predicted torques, but also fails to generalize the force predictions to the new data.

\subsection{Closed-Loop Comparison}\label{sec:closed_loop_comparison}
To demonstrate the benefits of an accurate force and torque model, a second set of experiments presents a comparison of closed-loop simulation performance. 
Using the simulation setup explained in Section~\ref{sec:method}, a set of unseen trajectories~(Figure~\ref{fig:trajectories}) is flown in simulation and the resulting flight path is compared with the data obtained from executing the same set of trajectories on the real platform. 
As for the previous set of experiments, also this comparison is performed for models identified on the full training set, as well as a reduced training set to compare generalization performance. 
Additionally to the baselines already used in the previous experiments, this experiment also compares against RotorS~\cite{furrer2016rotors}. 
To do this, the standard model in RotorS is updated with the parameters identified from the real platform (i.e. mass, inertia, dimensions, lift and drag coefficients). 

Table~\ref{tab:closed_loop_comparison} illustrates the results of the closed-loop experiment.
For each trajectory, the accumulated positional error between the simulated flight and the data observed in the real world is reported. 
As can be seen, all models achieve similar performance for trajectories close to hover. 
The \textit{Fit} model exhibits increasing bias with higher speeds, with the error exceeding the worst performance of the proposed approach already at average speeds below $\SI{7}{\meter\per\second}$.
In contrast, the \textit{BEM} model is able to maintain competitive performance up to the fastest trajectories.
At low speeds, the \textit{PolyFit} baseline performs very well, but exhibits increasing bias for higher speeds, even resulting in a crash on the fastest trajectory. 
The RotorS baseline performs inferior on all trajectories, achieving results comparable to the \textit{Fit} baseline.
The proposed approach combining BEM with a learned residual term (\textit{BEM+NN}) achieves competitive performance on the slow trajectories and outperforms all baselines on the faster trajectories.
Compared to \textit{Fit+NN}, \textit{BEM+NN} achieves consistently better performance for fast maneuvers. 

\begin{figure}
    \centering
    \input{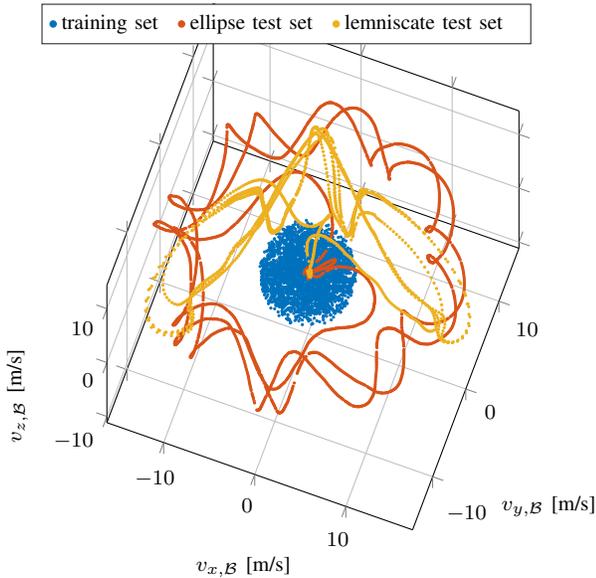}
    \caption{Visualization of the reduced training set, the ellipse test set and the fastest lemniscate test set in the body-frame velocity space. Although the test set mostly covers regions of the state space that are not part of the training set, the trained BEM+NN$^\star$ model still provides good accuracy in simulating the trajectory. This demonstrates its remarkable generalization capability.}
    \label{fig:generalization_V}
\end{figure}

When trained on the reduced dataset, all models show decreased performance. 
However, while approaches such as \textit{PolyFit} completely break down, the proposed approach experiences only a minor performance reduction around 20\%, outperforming the baselines on all faster trajectories. 
This result highlights the ability of the proposed approach to generalize beyond the data it was trained on (Figure~\ref{fig:generalization_V}). 

\section{Discussion}\label{sec:discussion}
The results obtained in this work show that the proposed hybrid dynamics model, combining first-principles based on blade-element-momentum theory with a learning-based residual term, outperforms state-of-the-art modeling for quadrotors with a 50\% decreased aerodynamic force and torque prediction error.
Furthermore, evaluation in controlled experiments on a large real-world dataset shows that such a complementary modeling approach outperforms each of its compositional submodules.

In fact, not only does the performance of the proposed hybrid model structure improve with a more capable rotor model, but the learned residual dynamics also increase in accuracy if a broader envelope of effects can already be captured using first principles. 
Specifically, the learned residual prediction achieves up to 30\% better performance when combined with a rotor model based on blade-element-momentum theory. 

While the proposed approach significantly improves upon state-of-the-art in quadrotor modeling for highly aggressive maneuvers by up to 60\%, its advantages for slow speed trajectories below \SI{5}{\meter\per\second} are limited.
Our experiments indicate that for such slow trajectories, a traditional parametric approach such as~\cite{sun2019quadrotor} achieves very strong performance at a lower computational cost. 
While the quadratic and polynomial fits can be evaluated in a mere \SI{1}{\micro\second}, even on a micro processor, the BEM model requires in the order of \rev{\SI{100}{\micro\second}} on a modern Intel-architecture CPU, and a forward pass of the network averages also at around \SI{100}{\micro\second} on a modern NVidia GPU.
The reason for the dominant runtime of the BEM model is the necessary implicit solution for  the induced velocity equation.
Even though our approach is not optimized for runtime, simulations can be run at an arbitrary timescale, where most applications gladly trade-off real-time evaluation for improved accuracy.

\rev{
The results of the closed-loop simulation using the proposed model could be further improved by refining the following aspects of the control pipeline:
(i)~The experimental platform currently relies on the BetaFlight inner-loop low-level controller, which is optimized for human pilots. However, as such, it only takes a throttle command and a body rate command as inputs, and relies on the inner-loop to track the rate command. Furthermore, it performs filtering and interpolation of the control signals to ensure a consistent flight feeling for human pilots, which introduces undesirable control-loop shaping. An MPC outer-loop controller directly outputting single-rotor motor speeds that are tracked by an inner loop motor speed controller would improve the accuracy of our approach further as this would minimize the differences between the simulation and the actual experiments. 
(ii)~Modeling the latency from the motion capture pose filtering, the data transmission to the drone, and the communication to the flight controller in simulation would also reduce the error as it improves the realism of the simulator. The authors expect the results in Table~\ref{tab:closed_loop_comparison} to be even more favorable for their approach in such an ideal setting. The accuracy of the force and torque predictions shown in Table~\ref{tab:forces_torques} would also benefit from a custom low-level controller providing more precise and less noisy motor-speed information.}

Compared to a purely learning-based approach such as \textit{None+NN}, the proposed approach performs 25\% better for all non-trivial trajectories with average speeds above $\SI{4}{\meter\per\second}$, and extrapolates well to unseen flight data, as opposed to e.g. the \textit{PolyFit} baseline.
The authors expect the performance of learning-only approaches to improve with more data.
However, in a real world setting, where high-quality data is sparse, pure learning-based approaches fall short of traditional methods.
Additionally, accurate aerodynamic force and torque prediction does not necessarily translate to good closed-loop performance, as demonstrated in our evaluation.
Moreover, this study observed cases where a purely-learned residual component introduced a feedback loop on the predicted torques that led to a crash. 
In such cases, support through first-principles is vital for accurate and robust modeling.

\section{Conclusion}\label{sec:conclusion}

This work proposes a novel method to model quadrotors by combining modeling based on first principles with a learning-based residual term represented by a neural network.
The proposed method is able to accurately model quadrotors even throughout aggressive trajectories pushing the platform to its limits. 
This hybrid model outperforms its compositional modules with up to 50\% error reduction, including baseline methods that utilize only first-principles modeling, as well as purely learning-based methods.
The method shows strong generalization beyond the training set used to identify the model and predicts accurate forces and torques where other methods break down. 
Controlled experiments indicate that the fusion of learned dynamics with first-principles is a powerful combination, where the learned dynamics-residual benefits from high-fidelity models, such as the BEM.
Applied to simulations, our approach enables unprecedented accuracy,
reducing positional RMSE from $\sim$\SI{0.8}{\meter} for state-of-the-art approaches, down to below \SI{0.3}{\meter}.
This could tremendously speed up development and testing of advanced control and navigation strategies for quadrotors, without the need of the tedious and crash-prone trial-and-error strategy on real systems.

\section*{Acknowledgement}
This work was supported by the National Centre of Competence in Research (NCCR) Robotics through the Swiss National Science Foundation (SNSF), the Intel Network on Intelligent
Systems, the European Union’s Horizon 2020 Research and Innovation Programme under grant agreement No. 871479 (AERIAL-CORE) and the European Research Council (ERC) under grant agreement No. 864042 (AGILEFLIGHT).

{\footnotesize
\balance
\bibliographystyle{unsrtnat}
\bibliography{references}

\begin{thebibliography}{41}
\providecommand{\natexlab}[1]{#1}
\providecommand{\url}[1]{\texttt{#1}}
\expandafter\ifx\csname urlstyle\endcsname\relax
  \providecommand{\doi}[1]{doi: #1}\else
  \providecommand{\doi}{doi: \begingroup \urlstyle{rm}\Url}\fi

\bibitem[Ryou et~al.(2020)Ryou, Tal, and Karaman]{ryou2020multi}
Gilhyun Ryou, Ezra Tal, and Sertac Karaman.
\newblock Multi-fidelity black-box optimization for time-optimal quadrotor
  maneuvers.
\newblock In \emph{Robotics: Science and Systems (RSS)}, 2020.

\bibitem[{Loianno} et~al.(2017){Loianno}, {Brunner}, {McGrath}, and
  {Kumar}]{loianno2017estimation}
G.~{Loianno}, C.~{Brunner}, G.~{McGrath}, and V.~{Kumar}.
\newblock Estimation, control, and planning for aggressive flight with a small
  quadrotor with a single camera and imu.
\newblock \emph{IEEE Robotics and Automation Letters}, 2\penalty0 (2):\penalty0
  404--411, 2017.
\newblock \doi{10.1109/LRA.2016.2633290}.

\bibitem[Kaufmann et~al.(2020)Kaufmann, Loquercio, Ranftl, M{\"u}ller, Koltun,
  and Scaramuzza]{kaufmann2020RSS}
Elia Kaufmann, Antonio Loquercio, Ren{\'e} Ranftl, Matthias M{\"u}ller, Vladlen
  Koltun, and Davide Scaramuzza.
\newblock Deep drone acrobatics.
\newblock \emph{RSS: Robotics, Science, and Systems}, 2020.

\bibitem[Foehn et~al.(2021)Foehn, Romero, and Scaramuzza]{foehn2021time}
Philipp Foehn, Angel Romero, and Davide Scaramuzza.
\newblock Time-optimal planning for quadrotor waypoint flight.
\newblock \emph{Science Robotics}, 2021.

\bibitem[Moon et~al.(2019)Moon, Martinez-Carranza, Cieslewski, Faessler,
  Falanga, Simovic, Scaramuzza, Li, Ozo, De~Wagter, et~al.]{moon2019challenges}
Hyungpil Moon, Jose Martinez-Carranza, Titus Cieslewski, Matthias Faessler,
  Davide Falanga, Alessandro Simovic, Davide Scaramuzza, Shuo Li, Michael Ozo,
  Christophe De~Wagter, et~al.
\newblock Challenges and implemented technologies used in autonomous drone
  racing.
\newblock \emph{Intelligent Service Robotics}, 12\penalty0 (2):\penalty0
  137--148, 2019.

\bibitem[Madaan et~al.(2020)Madaan, Gyde, Vemprala, Brown, Nagami, Taubner,
  Cristofalo, Scaramuzza, Schwager, and Kapoor]{madaan2019gameofdrones}
R.~Madaan, N.~Gyde, S.~Vemprala, M.~Brown, K.~Nagami, T.~Taubner,
  E.~Cristofalo, D.~Scaramuzza, M.~Schwager, and A.~Kapoor.
\newblock Airsim drone racing lab.
\newblock In \emph{PLMR Post Proceedings of the NeurIPS 2019 Competition
  Track}, 2020.

\bibitem[Foehn et~al.(2020)Foehn, Brescianini, Kaufmann, Cieslewski, Gehrig,
  Muglikar, and Scaramuzza]{foehn2020alphapilot}
Philipp Foehn, Dario Brescianini, Elia Kaufmann, Titus Cieslewski, Mathias
  Gehrig, Manasi Muglikar, and Davide Scaramuzza.
\newblock Alphapilot: Autonomous drone racing.
\newblock \emph{RSS: Robotics, Science, and Systems}, 2020.

\bibitem[Punjani and Abbeel(2015)]{punjani2015deep}
Ali Punjani and Pieter Abbeel.
\newblock Deep learning helicopter dynamics models.
\newblock In \emph{2015 IEEE International Conference on Robotics and
  Automation (ICRA)}, pages 3223--3230. IEEE, 2015.

\bibitem[Bansal et~al.(2016)Bansal, Akametalu, Jiang, Laine, and
  Tomlin]{bansal2016learning}
Somil Bansal, Anayo~K Akametalu, Frank~J Jiang, Forrest Laine, and Claire~J
  Tomlin.
\newblock Learning quadrotor dynamics using neural network for flight control.
\newblock In \emph{2016 IEEE 55th Conference on Decision and Control (CDC)},
  pages 4653--4660. IEEE, 2016.

\bibitem[Furrer et~al.(2016)Furrer, Burri, Achtelik, and
  Siegwart]{furrer2016rotors}
Fadri Furrer, Michael Burri, Markus Achtelik, and Roland Siegwart.
\newblock Rotors—a modular gazebo mav simulator framework.
\newblock In \emph{Robot Operating System (ROS)}, pages 595--625. Springer,
  2016.

\bibitem[Shah et~al.(2018)Shah, Dey, Lovett, and Kapoor]{shah2018airsim}
Shital Shah, Debadeepta Dey, Chris Lovett, and Ashish Kapoor.
\newblock Airsim: High-fidelity visual and physical simulation for autonomous
  vehicles.
\newblock In \emph{Field and service robotics}, pages 621--635. Springer, 2018.

\bibitem[Faessler et~al.(2017)Faessler, Franchi, and
  Scaramuzza]{faessler2017differential}
Matthias Faessler, Antonio Franchi, and Davide Scaramuzza.
\newblock Differential flatness of quadrotor dynamics subject to rotor drag for
  accurate tracking of high-speed trajectories.
\newblock \emph{{IEEE} Robot. Autom. Lett.}, 2017.

\bibitem[Sun et~al.(2019)Sun, de~Visser, and Chu]{sun2019quadrotor}
Sihao Sun, Coen~C de~Visser, and Qiping Chu.
\newblock Quadrotor gray-box model identification from high-speed flight data.
\newblock \emph{Journal of Aircraft}, 56\penalty0 (2):\penalty0 645--661, 2019.

\bibitem[Ventura~Diaz and Yoon(2018)]{ventura2018high}
Patricia Ventura~Diaz and Steven Yoon.
\newblock High-fidelity computational aerodynamics of multi-rotor unmanned
  aerial vehicles.
\newblock In \emph{2018 AIAA Aerospace Sciences Meeting}, page 1266, 2018.

\bibitem[Mohajerin et~al.(2018)Mohajerin, Mozifian, and
  Waslander]{mohajerin2018deep}
Nima Mohajerin, Melissa Mozifian, and Steven Waslander.
\newblock Deep learning a quadrotor dynamic model for multi-step prediction.
\newblock In \emph{2018 IEEE International Conference on Robotics and
  Automation (ICRA)}, pages 2454--2459. IEEE, 2018.

\bibitem[Portwood et~al.(2019)Portwood, Mitra, Ribeiro, Nguyen, Nadiga, Saenz,
  Chertkov, Garg, Anandkumar, and Dengel]{portwood2019turbulence}
Gavin~D Portwood, Peetak~P Mitra, Mateus~Dias Ribeiro, Tan~Minh Nguyen,
  Balasubramanya~T Nadiga, Juan~A Saenz, Michael Chertkov, Animesh Garg, Anima
  Anandkumar, and Andreas Dengel.
\newblock Turbulence forecasting via neural ode.
\newblock \emph{2nd Workshop on Machine Learning and the Physical Sciences
  (NeurIPS 2019)}, 2019.

\bibitem[Grzeszczuk et~al.(1998)Grzeszczuk, Terzopoulos, and
  Hinton]{grzeszczuk1998neuroanimator}
Radek Grzeszczuk, Demetri Terzopoulos, and Geoffrey Hinton.
\newblock Neuroanimator: Fast neural network emulation and control of
  physics-based models.
\newblock In \emph{Proceedings of the 25th annual conference on Computer
  graphics and interactive techniques}, pages 9--20, 1998.

\bibitem[Mahony et~al.(2012)Mahony, Kumar, and Corke]{mahony2012multirotor}
Robert Mahony, Vijay Kumar, and Peter Corke.
\newblock Multirotor aerial vehicles: Modeling, estimation, and control of
  quadrotor.
\newblock \emph{IEEE Robotics and Automation magazine}, 19\penalty0
  (3):\penalty0 20--32, 2012.

\bibitem[Song et~al.(2020)Song, Naji, Kaufmann, Loquercio, and
  Scaramuzza]{song2020flightmare}
Yunlong Song, Selim Naji, Elia Kaufmann, Antonio Loquercio, and Davide
  Scaramuzza.
\newblock Flightmare: A flexible quadrotor simulator.
\newblock 2020.

\bibitem[Meyer et~al.(2012)Meyer, Sendobry, Kohlbrecher, Klingauf, and
  Von~Stryk]{meyer2012comprehensive}
Johannes Meyer, Alexander Sendobry, Stefan Kohlbrecher, Uwe Klingauf, and Oskar
  Von~Stryk.
\newblock Comprehensive simulation of quadrotor uavs using ros and gazebo.
\newblock In \emph{International conference on simulation, modeling, and
  programming for autonomous robots}, pages 400--411. Springer, 2012.

\bibitem[Hoffmann et~al.(2007)Hoffmann, Huang, Waslander, and
  Tomlin]{hoffmann2007quadrotor}
Gabriel Hoffmann, Haomiao Huang, Steven Waslander, and Claire Tomlin.
\newblock Quadrotor helicopter flight dynamics and control: Theory and
  experiment.
\newblock In \emph{AIAA guidance, navigation and control conference and
  exhibit}, page 6461, 2007.

\bibitem[Huang et~al.(2009)Huang, Hoffmann, Waslander, and
  Tomlin]{huang2009aerodynamics}
Haomiao Huang, Gabriel~M Hoffmann, Steven~L Waslander, and Claire~J Tomlin.
\newblock Aerodynamics and control of autonomous quadrotor helicopters in
  aggressive maneuvering.
\newblock In \emph{2009 IEEE international conference on robotics and
  automation}, pages 3277--3282. IEEE, 2009.

\bibitem[Hoffmann et~al.(2011)Hoffmann, Huang, Waslander, and
  Tomlin]{hoffmann2011precision}
Gabriel~M Hoffmann, Haomiao Huang, Steven~L Waslander, and Claire~J Tomlin.
\newblock Precision flight control for a multi-vehicle quadrotor helicopter
  testbed.
\newblock \emph{Control engineering practice}, 19\penalty0 (9):\penalty0
  1023--1036, 2011.

\bibitem[Prouty(1995)]{prouty1995helicopter}
Raymond~W Prouty.
\newblock \emph{Helicopter performance, stability, and control}.
\newblock 1995.

\bibitem[Orsag and Bogdan(2012)]{orsag2012influence}
Matko Orsag and Stjepan Bogdan.
\newblock Influence of forward and descent flight on quadrotor dynamics.
\newblock \emph{Recent Advances in Aircraft Technology}, pages 141--156, 2012.

\bibitem[Bristeau et~al.(2009)Bristeau, Martin, Sala{\"u}n, and
  Petit]{bristeau2009role}
Pierre-Jean Bristeau, Philippe Martin, Erwan Sala{\"u}n, and Nicolas Petit.
\newblock The role of propeller aerodynamics in the model of a quadrotor uav.
\newblock In \emph{2009 European control conference (ECC)}, pages 683--688.
  IEEE, 2009.

\bibitem[Kaya and Kutay(2014)]{kaya2014aerodynamic}
Derya Kaya and Ali~T Kutay.
\newblock Aerodynamic modeling and parameter estimation of a quadrotor
  helicopter.
\newblock In \emph{AIAA Atmospheric Flight Mechanics Conference}, page 2558,
  2014.

\bibitem[Tang and Li(2015)]{tang2015dynamic}
Yi-Rui Tang and Yangmin Li.
\newblock Dynamic modeling for high-performance controller design of a uav
  quadrotor.
\newblock In \emph{2015 IEEE International Conference on Information and
  Automation}, pages 3112--3117. IEEE, 2015.

\bibitem[Powers et~al.(2013)Powers, Mellinger, Kushleyev, Kothmann, and
  Kumar]{powers2013influence}
Caitlin Powers, Daniel Mellinger, Aleksandr Kushleyev, Bruce Kothmann, and
  Vijay Kumar.
\newblock Influence of aerodynamics and proximity effects in quadrotor flight.
\newblock In \emph{Experimental robotics}, pages 289--302. Springer, 2013.

\bibitem[Khan and Nahon(2013)]{khan2013toward}
Waqas Khan and Meyer Nahon.
\newblock Toward an accurate physics-based uav thruster model.
\newblock \emph{IEEE/ASME Transactions on Mechatronics}, 18\penalty0
  (4):\penalty0 1269--1279, 2013.

\bibitem[Gill and D'Andrea(2017)]{gill2017propeller}
Rajan Gill and Raffaello D'Andrea.
\newblock Propeller thrust and drag in forward flight.
\newblock In \emph{2017 IEEE Conference on Control Technology and Applications
  (CCTA)}, pages 73--79. IEEE, 2017.

\bibitem[Gill and D’Andrea(2019)]{gill2019computationally}
Rajan Gill and Raffaello D’Andrea.
\newblock Computationally efficient force and moment models for propellers in
  uav forward flight applications.
\newblock \emph{Drones}, 3\penalty0 (4):\penalty0 77, 2019.

\bibitem[Russell et~al.(2016)Russell, Jung, Willink, and
  Glasner]{russell2016wind}
Carl Russell, Jaewoo Jung, Gina Willink, and Brett Glasner.
\newblock Wind tunnel and hover performance test results for multicopter uas
  vehicles.
\newblock In \emph{AHS 72nd annual forum}, pages 16--19, 2016.

\bibitem[Schiano et~al.(2014)Schiano, Alonso-Mora, Rudin, Beardsley, Siegwart,
  and Sicilianok]{schiano2014towards}
Fabrizio Schiano, Javier Alonso-Mora, Konrad Rudin, Paul Beardsley, Roland~Y
  Siegwart, and Bruno Sicilianok.
\newblock Towards estimation and correction of wind effects on a quadrotor uav.
\newblock In \emph{IMAV 2014: International Micro Air Vehicle Conference and
  Competition 2014}, pages 134--141, 2014.

\bibitem[Baris et~al.(2019)Baris, Britcher, and Altamirano]{baris2019wind}
Engin Baris, Colin~P Britcher, and George Altamirano.
\newblock Wind tunnel testing of static and dynamic aerodynamic characteristics
  of a quadcopter.
\newblock In \emph{AIAA Aviation 2019 Forum}, page 2973, 2019.

\bibitem[Torrente et~al.(2021)Torrente, Kaufmann, F{\"o}hn, and
  Scaramuzza]{torrente2021data}
Guillem Torrente, Elia Kaufmann, Philipp F{\"o}hn, and Davide Scaramuzza.
\newblock Data-driven {MPC} for quadrotors.
\newblock \emph{{IEEE} Robot. Autom. Lett.}, 2021.

\bibitem[Luo et~al.(2015)Luo, Zhu, and Yan]{luo2015novel}
Jinglin Luo, Longfei Zhu, and Guirong Yan.
\newblock Novel quadrotor forward-flight model based on wake interference.
\newblock \emph{Aiaa Journal}, 53\penalty0 (12):\penalty0 3522--3533, 2015.

\bibitem[Mohajerin and Waslander(2019)]{mohajerin2019multistep}
Nima Mohajerin and Steven Waslander.
\newblock Multistep prediction of dynamic systems with recurrent neural
  networks.
\newblock \emph{IEEE Transactions on Neural Networks and Learning Systems},
  2019.

\bibitem[Shi et~al.(2019)Shi, Shi, O’Connell, Yu, Azizzadenesheli,
  Anandkumar, Yue, and Chung]{shi2019learnedlanding}
Guanya Shi, Xichen Shi, Michael O’Connell, Rose Yu, Kamyar Azizzadenesheli,
  Animashree Anandkumar, Yisong Yue, and Soon-Jo Chung.
\newblock Neural lander: Stable drone landing control using learned dynamics.
\newblock \emph{2019 International Conference on Robotics and Automation
  (ICRA)}, May 2019.
\newblock \doi{10.1109/icra.2019.8794351}.
\newblock URL \url{http://dx.doi.org/10.1109/ICRA.2019.8794351}.

\bibitem[Ducard and Hua(2014)]{ducard2014modeling}
Guillaume Ducard and Minh-Duc Hua.
\newblock Modeling of an unmanned hybrid aerial vehicle.
\newblock In \emph{2014 IEEE Conference on Control Applications (CCA)}, pages
  1011--1016. IEEE, 2014.

\bibitem[Oord et~al.(2016)Oord, Dieleman, Zen, Simonyan, Vinyals, Graves,
  Kalchbrenner, Senior, and Kavukcuoglu]{oord2016wavenet}
Aaron van~den Oord, Sander Dieleman, Heiga Zen, Karen Simonyan, Oriol Vinyals,
  Alex Graves, Nal Kalchbrenner, Andrew Senior, and Koray Kavukcuoglu.
\newblock Wavenet: A generative model for raw audio.
\newblock \emph{arXiv preprint arXiv:1609.03499}, 2016.

\end{thebibliography}
}

\end{document}